\documentclass[jair,twoside,11pt,theapa]{article}
\usepackage{jair, theapa, rawfonts}
\usepackage[utf8]{inputenc}

\usepackage{graphicx}
\usepackage{xcolor}
\usepackage{xspace}
\usepackage{xstring}
\usepackage{tabularx}
\usepackage{multirow}
\usepackage{tikz-dependency}
\usepackage{covington}

\usepackage{amsmath}
\usepackage{amssymb}
\usepackage{mathtools}

\usepackage{soul}
\usepackage{tabu}

\usepackage{bm}
\usepackage{paralist}
\usepackage{tablefootnote}
\usepackage{subfig}
\usepackage{capt-of}
\usepackage{scalefnt}

\usepackage{latexsym}
\usepackage{paralist}

\usepackage{calc}
\usepackage{booktabs}
\usepackage{url}

\usepackage{scrextend}

\newcommand{\R}{\mathbb{R}}
\DeclareMathOperator*{\argmin}{arg\,min}

\newcommand{\rt}[1]{\rotatebox{90}{#1}}

\newcommand{\ie}{\textit{i.\,e.}\xspace}
\newcommand{\eg}{\textit{e.\,g.}\xspace}
\newcommand{\ia}{\textit{i.\,a.}\xspace}

\newcommand{\F}{$\text{F}_1$\xspace}

\newcommand{\lstms}{\textsc{Lstms}\xspace}

\newcommand{\cnns}{\textsc{Cnns}\xspace}

\newcommand{\blse}{\textsc{Blse}\xspace}
\newcommand{\mt}{\textsc{MT}\xspace}
\newcommand{\barista}{\textsc{Barista}\xspace}
\newcommand{\vecmap}{\textsc{VecMap}\xspace}
\newcommand{\unsup}{\textsc{Unsup}\xspace}
\newcommand{\mono}{\textsc{Mono}\xspace}
\newcommand{\ensemble}{\textsc{Ensemble}\xspace}
\newcommand{\muse}{\textsc{Muse}\xspace}
\newcommand{\splitsent}{\textsc{Split}\xspace}
\newcommand{\sent}{\textsc{Sent}\xspace}
\newcommand{\cononly}{\textsc{Context-only}\xspace}
\newcommand{\asponly}{\textsc{Target-only}\xspace}
\newcommand{\llcon}{\textrm{con}_\ell}
\newcommand{\rcon}{\textrm{con}_r}
\newcommand{\cononlytable}{\textsc{\shortstack[1]{Context-\\only}}}
\newcommand{\asponlytable}{\textsc{\shortstack[1]{Target-\\only}}}

\newcommand{\Csource}{C_{\textrm{source}}}
\newcommand{\Ctarget}{C_{\textrm{target}}}
\newcommand{\mathN}{\mathcal{N}}
\newcommand{\ccc}[1]{\multicolumn{1}{c}{#1}}

\newcommand\footnoteref[1]{\protected@xdef\@thefnmark{\ref{#1}}\@footnotemark}
\newcommand{\aspectbox}[1]{{\setlength{\fboxsep}{1pt}\colorbox{lightgreen}{#1}}}
\newcommand{\posbox}[1]{{\setlength{\fboxsep}{1pt}\colorbox{lightblue}{#1}}}
\newcommand{\negbox}[1]{{\setlength{\fboxsep}{1pt}\colorbox{lightred}{#1}}}

\definecolor{lightgreen}{RGB}{200,255,200}
\definecolor{lightblue}{RGB}{200,200,255}
\definecolor{lightred}{RGB}{255,200,200}

\ShortHeadings{Targeted Cross-Lingual Sentiment Analysis}
{Barnes, Klinger}
\firstpageno{1}

\begin{document}

\title{%
Embedding Projection for Targeted Cross-Lingual Sentiment:\\ Model Comparisons and a Real-World Study
}

\author{\name Jeremy Barnes \email jeremycb@ifi.uio.no \\
       \addr Language Technology Group\\
       University of Oslo \\
       Gaustadalléen 23 B, N-0373\\
       Oslo, Norway
       \AND
       \name Roman Klinger \email klinger@ims.uni-stuttgart.de \\
       \addr Institut für Maschinelle Sprachverarbeitung\\
       University of Stuttgart \\
       Pfaffenwaldring 5b, 70569\\
       Stuttgart, Germany}

\maketitle

\begin{abstract}\noindent
  Sentiment analysis benefits from large, hand-annotated resources in
  order to train and test machine learning models, which are often
  data hungry. While some languages, \eg, English, have a vast array
  of these resources, most under-resourced languages do not,
  especially for fine-grained sentiment tasks, such as aspect-level or
  targeted sentiment analysis. To improve this situation, we propose a
  \emph{cross-lingual} approach to sentiment analysis that is
  applicable to under-resourced languages and takes into account
  target-level information. This model incorporates sentiment
  information into bilingual distributional representations, by
  jointly optimizing them for semantics and sentiment, showing
  state-of-the-art performance at sentence-level when combined with
  machine translation. The adaptation to targeted sentiment analysis
  on multiple domains shows that our model outperforms other
  projection-based bilingual embedding methods on binary targeted
  sentiment tasks. Our analysis on ten languages demonstrates that the
  amount of unlabeled monolingual data has surprisingly little
  effect on the sentiment results. As expected, the choice of a
  annotated source language for projection to a target leads to better
  results for source-target language pairs which are
  similar. Therefore, our results suggest that more efforts should be
  spent on the creation of resources for less similar languages to
  those which are resource-rich already. Finally, a domain mismatch
  leads to a decreased performance. This suggests resources in any
  language should ideally cover varieties of domains.
\end{abstract}

\section{Introduction}
\subsection{Targeted Sentiment Classification}
Opinions are everywhere in our lives. Every time we open a book, read
the newspaper, or look at social media, we scan for opinions or form
them ourselves. We are cued to the opinions of others, and often use
this information to update our own opinions
\shortcite{Asch1955,Das2014}. This is true on the Internet as much as
it is in our face-to-face relationships. In fact, with its wealth of
opinionated material available online, it has become feasible and
interesting to harness this data in order to automatically identify
opinions, which had previously been far more expensive and tedious when the only
access to data was offline.

\emph{Sentiment analysis}, sometimes referred to as \emph{opinion
  mining}, seeks to create data-driven methods to classify the
polarity of a text. The information obtained from sentiment
classifiers can then be used for tracking user opinions in different
domains \shortcite{Pang2002,Socher2013b,Nakov2013}, predicting the outcome
of political elections \shortcite{wang2012demo,bakliwal2013}, detecting
hate speech online \shortcite{Nahar2012,hartung-EtAl:2017:WASSA2017}, as
well as predicting changes in the stock market \shortcite{Pogolu2016}.

Sentiment analysis can be modeled as a classification task, especially
at sentence- and document-level, or as a sequence-labeling task at
target-level. Targeted sentiment analysis aims at predicting the
polarity expressed towards a particular entity or sub-aspect of that
entity. This is a more realistic view of sentiment, as polarities are
directed towards targets, not spread uniformly across sentences or
documents. Take the following example, where we mark the sentiment
target with \aspectbox{green}, positive sentiment expressions with
\posbox{blue}, and negative sentiment expressions with \negbox{red}.:
\begin{quote}
  The \aspectbox{café} near my house has \posbox{great}
  \aspectbox{coffee} but I\\ never go there because the
  \aspectbox{service} is \negbox{terrible}.
\end{quote}
In this sentence, it is
not stated what the sentiment towards the target ``café'' is, while
the sentiment of the target ``coffee'' is positive and that of
``service'' is negative. In order to correctly classify the sentiment
of each target, it is necessary to (1) detect the targets, (2)~detect
polarity expressions, and (3) resolve the relations between these. 

In order to model these relationships and test the accuracy of the
learned models, hand-annotated resources are typically used for
training machine learning algorithms. Resource-rich languages, \eg,
English, have high-quality annotated data for both classification and
sequence-labeling tasks, as well as for a variety of domains. However,
under-resourced languages either completely lack annotated data or
have only a few resources for specific domains or sentiment tasks. For
instance, for aspect-level sentiment analysis, English has datasets
available in the news domain \shortcite{Wiebe2005}, product review
domain \shortcite{HuandLiu2004,Ding2008,Pontiki2014,Pontiki2015},
education domain \shortcite{Welch2016}, medical domain
\shortcite{Grasser2018}, urban neighborhood domain
\shortcite{Saeidi2016}, and financial \shortcite{Maia2018}
domain. Spanish, on the other hand, has only three datasets
\shortcite{Agerri2013,Pontiki2016}, while Basque and Catalan only have
one each for a single domain \shortcite{Barnes2018a}. The cost of
annotating data can often be prohibitive as training native-speakers
to annotate fine-grained sentiment is a long process. This motivates
the need to develop sentiment analysis methods capable of leveraging
data annotated in other languages.

\subsection{Cross-Lingual Approaches to Sentiment Analysis}
Previous work on \emph{cross-lingual sentiment analysis} (CLSA) offers
a way to perform sentiment analysis in an under-resourced language
that does not have any annotated data available. Most methods relied
on the availability of large amounts of parallel data to transfer
sentiment information across languages. \emph{Machine translation}
(\mt), for example, has been the most common approach to cross-lingual
sentiment analysis
\shortcite{Banea2013,Almeida2015,Zhang2017}. Machine translation,
however, can be biased towards domains
\shortcite{Hua2008,Bertoldi2009,Koehn2017}, does not always preserve
sentiment \shortcite{Mohammad2016}, and requires millions of parallel
sentences \shortcite{Gavrila2011,Vaswani2017}, which places a limit on
which languages can benefit from these approaches. The following
example illustrates that \mt does not preserve sentiment (hotel review
in Basque, automatically translated via \url{translate.google.com}):
\begin{quote}
\aspectbox{Hotel}$^{1}$ \posbox{txukuna} da, \posbox{nahiko berria}. \aspectbox{Harreran zeuden langileen arreta}$^{2}$ \negbox{ez zen onena} izan. \aspectbox{Tren geltoki bat}$^{3}$ du \posbox{5 minutura} eta \aspectbox{kotxez}$^{4}$ \posbox{berehala iristen} da baina \aspectbox{oinez}$^{5}$ \negbox{urruti samar} dago.\\

\aspectbox{The hotel}$^{1}$ is \posbox{tidy}, \posbox{quite new}. \aspectbox{The care of the workers at reception}$^{2}$ was \negbox{not the best}. It's \posbox{5 minutes away} from \aspectbox{a train station}$^{3}$ and it's \posbox{quick to reach} \aspectbox{the car}$^{4}$, but it's \posbox{a short distance away}.
\label{example:basque:unannotated}
\end{quote}
While the first two sentences are mostly well translated for the
purposes of sentiment analysis, in the third, there are a number of
reformulations and deletions that lead to a loss of information. It
should read ``It has a train station five minutes away and by car you
can reach it quickly, but by foot it's quite a distance.'' We can see
that one of the targets has been deleted and the sentiment has flipped
from negative to positive. Such common problems degrade the results of
cross-lingual sentiment systems that use \mt, especially at
target-level.

Although high quality machine translation systems exist between many
languages and have been shown to enable cross-lingual sentiment
analysis, for the vast majority of language pairs in the world there
is not enough parallel data to create these high quality \mt
systems. This lack of parallel data coupled with the computational
expense of \mt means that approaches to cross-lingual sentiment
analysis that do not require \mt should be preferred. Additionally,
most cross-lingual sentiment approaches using \mt have concentrated on
sentence- and document-level, and have not explored targeted or
aspect-level sentiment tasks.

\subsection{Bilingual Distributional Models and the Contributions of this Paper}
Recently, several \emph{bilingual distributional semantics models}
(bilingual embeddings) have been proposed and provide a useful
framework for cross-lingual research without requiring machine
translation. They are effective at generating features for bilingual
dictionary induction
\shortcite{Mikolov2013translation,Artetxe2016,Lample2017},
cross-lingual text classification
\shortcite{Prettenhofer2011b,Chandar2014}, or cross-lingual dependency
parsing \shortcite{Sogaard2015}, among others. In this framework,
words are represented as $n$-dimensional vectors which are created on
large monolingual corpora in order to (1)~maximize the similarity of
words that appear in similar contexts and use some bilingual
regularization in order to (2) maximize the similarity of translation
pairs. In this work, we concentrate on a subset of these bilingual
embedding methods that perform a post-hoc mapping to a bilingual
space, which we refer to as \emph{embedding projection methods}. One
of the main advantages of these methods is that they make better use
of small amounts of parallel data than \mt systems, even enabling
unsupervised machine translation \shortcite{Artetxe2018,Lample2018}.

With this paper, we provide the first extensive evaluation of
cross-lingual embeddings for targeted sentiment tasks. We formulate
the task of targeted sentiment analysis as classification, given the
targets from an oracle\footnote{This is a common assumption when
  studying target-level sentiment analysis
  \shortcite{Dong2014,Zhang2016}.}. The question we attempt to address
is \emph{how to infer the polarity of a sentiment target in a language
  that does not have any annotated sentiment data or parallel corpora
  with a resource-rich language.} In the following Catalan sentence,
for example, how can we determine that the sentiment of ``servei'' is
negative, while that of ``menjar'' is positive if we do not have
annotated data in Catalan or parallel data for English-Catalan?
\begin{quote}
  El \aspectbox{servei} al \aspectbox{restaurant} va ser
  \negbox{péssim}. Al menys el \aspectbox{menjar} era
  \posbox{bo}.
\end{quote}
Specifically, we propose an approach which requires (1) minimal
bilingual data and instead makes use of (2) high-quality monolingual
word embeddings in the source and target language. We take an
intermediate step by first testing this approach on sentence-level
classification. After confirming that our approach performs well at
sentence-level, we propose a targeted model with the same data
requirements. The main contributions are that we

\begin{itemize}
\item compare projection-based cross-lingual methods to \mt,
\item extend previous cross-lingual approaches
  to enable targeted cross-lingual sentiment analysis
  with minimal parallel data requirements,
\item compare different model architectures for cross-lingual
  targeted sentiment analysis,
\item perform a detailed error analysis, and detailing the advantages
  and disadvantages of each method,
\item and, finally, deploy the methods in a realistic case-study to
  analyze their suitability beyond applications on (naturally) limited
  language pairs.
\end{itemize}

In addition, we make our code and data publicly available at
\url{https://github.com/jbarnesspain/targeted_blse} to support future
research. The rest of the article is organized as follows: In
Section~\ref{sec:previouswork}, we detail related work and motivate
the need for a different approach. In Section~\ref{sec:projecting}, we
describe both the sentence-level and targeted projection approaches
that we propose. In Section~\ref{sec:experiments}, we detail the
resources and experimental setup for both sentence and targeted
classification. In Section~\ref{sec:results}, we describe the results
of the two experiments, as well as perform a detailed error
analysis. In Section~\ref{sec:deployment}, we perform a case study
whose purpose is to give a more qualitative view of the
models. Finally, we discuss the implications of the results in
Section~\ref{sec:conclusion}.

\section{Previous Work}
\label{sec:previouswork}
\noindent Sentiment analysis has become an enormously popular task
with a focus on classification approaches on individual languages, but
there has not been as much work on cross-lingual approaches. In this
section, we detail the most relevant work on cross-lingual sentiment
analysis and lay the basis for the bilingual embedding approach we
propose later.

\subsection{Machine Translation Based Methods}
Early work in cross-lingual sentiment
analysis found that machine translation (\mt) had reached a point of
maturity that enabled the transfer of sentiment across languages.
Researchers translated sentiment lexicons \shortcite{Mihalcea2007,Meng2012}
or annotated corpora and used word alignments to project sentiment
annotation and create target-language annotated corpora
\shortcite{Banea2008,Duh2011a,Demirtas2013,Balahur2014d}.

Several approaches included a multi-view representation of the data
\shortcite{Banea2010,Xiao2012} or co-training \shortcite{Wan2009,Demirtas2013}
to improve over a naive implementation of machine
translation, where only the translated version of the data is considered. There are also approaches
which only require parallel data \shortcite{Meng2012,Zhou2016,Rasooli2017},
instead of machine translation.

All of these approaches, however, require large amounts of parallel
data or an existing high quality translation tool, which are not
always available. To tackle this issue, \shortciteA{Barnes2016}
explore cross-lingual approaches for aspect-based sentiment analysis,
comparing machine translation methods and those that instead rely on
bilingual vector representations. They conclude that \mt approaches
outperform current bilingual representation methods.

\shortciteA{Chen2016} propose an
adversarial deep averaging network, which trains a joint feature
extractor for two languages. They minimize the difference between
these features across languages by learning to fool a language
discriminator. This requires no parallel data, but does
require large amounts of unlabeled data and has not been tested on
fine-grained sentiment analysis.

\subsection{Bilingual Embedding Methods} 
Recently proposed bilingual
embedding methods \shortcite{Hermann2014,Chandar2014,Gouws2015} offer a
natural way to bridge the language gap. These particular approaches to
bilingual embeddings, however, also require large parallel corpora in order
to build the bilingual space, which gives no advantage over machine
translation. Another approach to creating bilingual word embeddings,
which we refer to as \emph{Projection-based Bilingual Embeddings},
has the advantage of requiring relatively
little parallel training data while taking advantage of larger amounts
of monolingual data. In the following, we describe the most
relevant approaches.

\paragraph{Bilingual Word Embedding Mappings (\vecmap):}
\shortciteA{Mikolov2013translation}
find that vector spaces in different languages have similar
arrangements. Therefore, they propose a linear projection which
consists of learning a rotation and scaling
matrix. \shortciteA{Artetxe2016,Artetxe2017} improve upon this approach
by requiring the projection to be orthogonal, thereby preserving the
monolingual quality of the original word vectors.

Given source embeddings $S$, target embeddings~$T$, and a bilingual
lexicon $L$, \shortciteA{Artetxe2016} learn a projection matrix $W$ by
minimizing the square of Euclidean distances
\begin{equation}
\argmin_W \sum_{i} ||S'W-T'||_{F}^{2}\,,
\end{equation}
where $S' \in S$ and $T' \in T$ are the word embedding matrices for
the tokens in the bilingual lexicon $L$. This is solved using the
Moore-Penrose pseudoinverse $S'^{+} = (S'^{T}S')^{-1}S'^{T}$ as $ W =
S'^{+}T'$, which can be computed using SVD. We refer to this approach
as \vecmap.

\paragraph{Multilingual Unsupervised and Supervised Embeddings
  (\muse)}

\shortciteA{Lample2017}
propose a similar refined orthogonal projection method to
\shortciteA{Artetxe2017}, but include an adversarial discriminator, which
seeks to discriminate samples from the projected space $WS$, and the
target $T$, while the projection matrix $W$ attempts to prevent this
making the projection from the source space $WS$ as similar to the
target space $T$ as possible.

They further refine their projection matrix by reducing the hubness
problem \shortcite{Dinu2015}, which is commonly found in high-dimensional
spaces. For each projected embedding $Wx$, they define the $k$ nearest
neighbors in the target space, $\mathN_{T}$, suggesting $k = 10$. They
consider the mean cosine similarity $r_{T}(Wx)$ between a projected embedding $Wx$
and its $k$ nearest neighbors
\begin{equation}
r_{T}(Wx) = \frac{1}{k} \sum_{y \in \mathN_{T}(Wx) } \cos(Wx,y)
\end{equation}
as well as the mean cosine of a target word $y$ to its neighborhood,
which they denote by $r_{S}$.

In order to decrease similarity between mapped vectors lying in dense
areas, they introduce a cross-domain similarity local scaling term
(CSLS)
\begin{equation}
\textrm{CSLS}(Wx,y) = 2 \cos(Wx,y) - r_{T}(Wx) - r_{S}(y)\,,
\end{equation}
which they find improves accuracy, while not requiring any parameter
tuning.

\paragraph{\barista}
\label{sec:barista}
\shortciteA{Gouws2015taskspecific} propose a method to create a
pseudo-bilingual corpus with a small task-specific bilingual lexicon,
which can then be used to train bilingual embeddings (\barista). This
approach requires a monolingual corpus in both the source and target
languages and a set of translation pairs. The source and target
corpora are concatenated and then every word is randomly kept or
replaced by its translation with a probability of 0.5. Any kind of
word embedding algorithm can be trained with this pseudo-bilingual
corpus to create bilingual word embeddings.

\subsection{Sentiment Embeddings}
 \shortciteA{Maas2011} first explored the
idea of incorporating sentiment information into semantic word
vectors. They proposed a topic modeling approach similar to latent
Dirichlet allocation in order to collect the semantic information in
their word vectors. To incorporate the sentiment information, they
included a second objective whereby they maximize the probability of
the sentiment label for each word in a labeled document.

\shortciteA{Tang2014} exploit distantly annotated tweets to
create Twitter sentiment embeddings. To incorporate distributional
information about tokens, they use a hinge loss and maximize the
likelihood of a true $n$-gram over a corrupted $n$-gram. They include a
second objective where they classify the polarity of the tweet given
the true $n$-gram. While these techniques have proven useful, they are
not easily transferred to a cross-lingual setting.

\shortciteA{Zhou2015} create bilingual sentiment embeddings by
translating all source data to the target language and vice versa. This 
requires the existence of a machine translation system, which is a
prohibitive assumption for many under-resourced languages, especially
if it must be open and freely accessible. This motivates approaches which can
use smaller amounts of parallel data to achieve similar results.

\subsection{Targeted Sentiment Analysis}
The methods discussed so far focus on classifying textual phrases like
documents or sentences. Next to these approaches, others have
concentrated on classifying aspects
\shortcite{HuandLiu2004,Liu2012,Pontiki2014} or targets
\shortcite{Zhang2015,Zhang2016,Tang2016} to assign them with polarity
values.

A common technique when adapting neural architectures to targeted
sentiment analysis is to break the text into left context, target, and
right context \shortcite{Zhang2015,Zhang2016}, alternatively keeping
the target as the final/beginning token in the respective contexts
\shortcite{Tang2016}. The model then extracts a feature vector from
each context and target, using some neural architecture, and
concatenates the outputs for classification.

More recent approaches attempt to augment a neural network with memory
to model these interactions
\shortcite{Chen2017,Xue2018,Wang2018,Liu2018}. \shortciteA{Wang2017}
explore methods to improve classification of multiple aspects in
tweets, while \shortciteA{Akhtar2018} attempt to use cross-lingual and
multilingual data to improve aspect-based sentiment analysis in
under-resourced languages.

As mentioned before, \mt has traditionally been the main approach for
transferring information across language barriers \cite[\ia, for
cross-lingual target-level sentiment analysis]{Klinger2015}. But this
is particularly problematic for targeted sentiment analysis, as
changes in word order or loss of words created during translation can
directly affect the performance of a classifier
\shortcite{Lambert2015}.

\section{Projecting Sentiment Across Languages}
\label{sec:projecting}

In this section, we propose a novel approach to incorporate sentiment
information into bilingual embeddings, which we first test on
\emph{sentence-level} cross-lingual sentiment
classification\footnote{This first contribution in this paper is an
  extended version of the work presented as \shortciteA{Barnes2018b}.}. We
then propose an extension in order to adapt this approach to
\emph{targeted} cross-lingual sentiment classification.  Our model,
\emph{Bilingual Sentiment Embeddings} (\blse), are embeddings that are
jointly optimized to represent both (a) semantic information in the
source and target languages, which are bound to each other through a
small bilingual dictionary, and (b) sentiment information, which is
annotated on the source language only. We only need three resources:
(1) a comparably small bilingual lexicon, (2) an annotated sentiment
corpus in the resource-rich language, and (3) monolingual word
embeddings for the two involved languages.

\subsection{Sentence-level Model}
\label{section:blse}

\begin{figure*}[]
\centering
\includegraphics[width=0.9\textwidth]{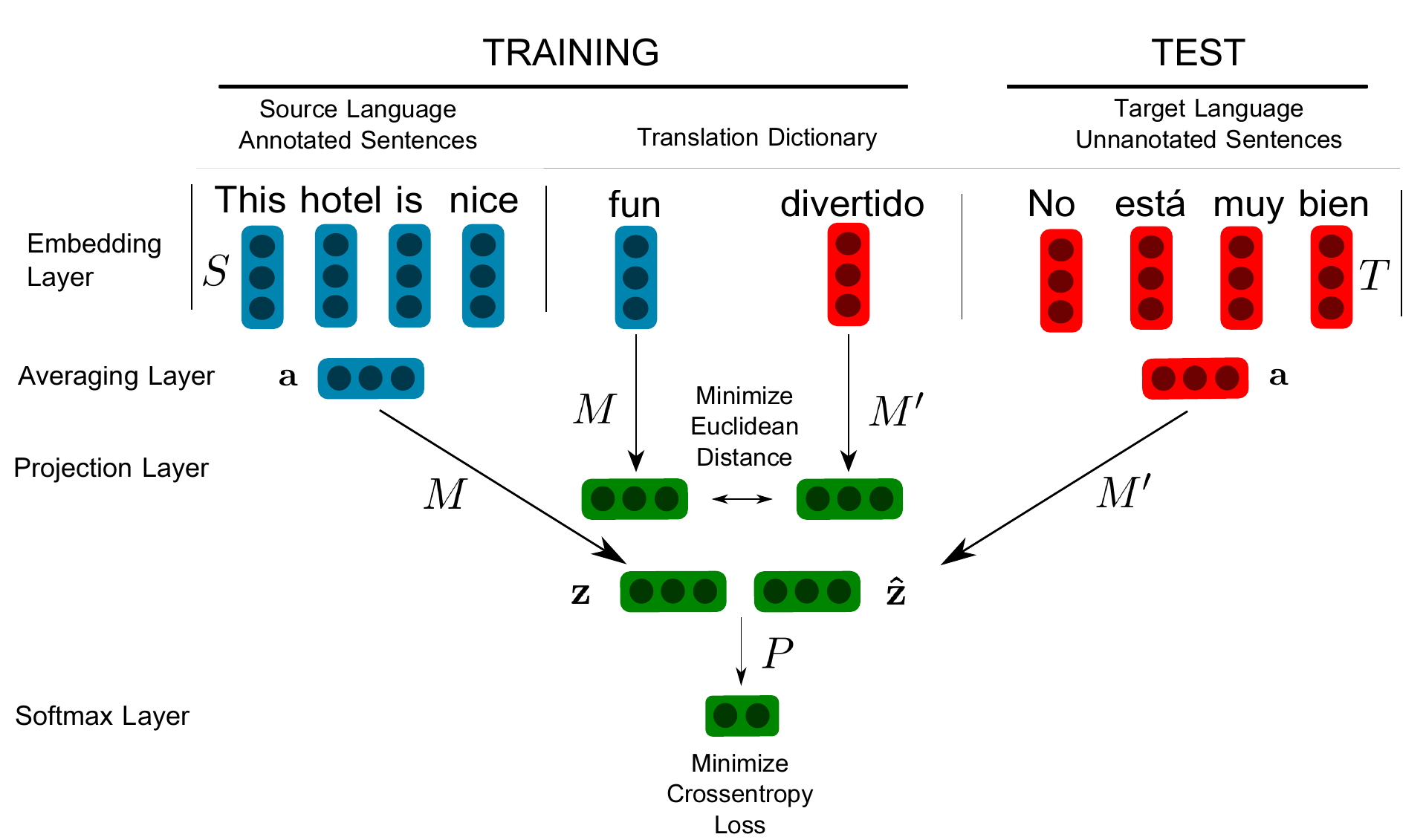}
\caption{Bilingual Sentiment Embedding Model (\blse)}
\label{acl:fig:model}
\end{figure*}

In this section, we detail the projection objective, the sentiment
objective, and finally the full objective for sentence-level
cross-lingual sentiment classification. A sketch of the full
sentence-level model is depicted in Figure~\ref{acl:fig:model}.

\subsubsection{Cross-lingual Projection}
\label{acl:crosslingual}

We assume that we have two precomputed vector spaces $S = \R^{v \times
  d}$ and $T = \R^{v' \times d'}$ for our source and target languages,
where $v$ ($v'$) is the length of the source vocabulary (target
vocabulary) and $d$ ($d'$) is the dimensionality of the embeddings.
We also assume that we have a bilingual lexicon $L$ of length $n$
which consists of word-to-word translation pairs $L$ = $\{(s_{1},t_{1}),
(s_{2},t_{2}),\ldots, (s_{n}, t_{n})\}$ which map from source to
target.

In order to create a mapping from both original vector spaces $S$ and
$T$ to shared sentiment-informed bilingual spaces $\mathbf{z}$ and
$\mathbf{\hat{z}}$, we employ two linear projection matrices, $M$ and
$M'$. During training, for each translation pair in $L$, we first look
up their associated vectors, project them through their associated
projection matrix and finally minimize the mean squared error of the
two projected vectors. This is similar to the approach taken by
\shortcite{Mikolov2013translation} , but includes an additional target
projection matrix.

The intuition for including this second matrix is that a single
projection matrix does not support the transfer of sentiment
information from the source language to the target language. Without
$M'$, any signal
coming from the sentiment classifier (see Section \ref{acl:sentiment}) would
have no affect on the target embedding space $T$, and optimizing
$M$ to predict sentiment and projection would only be
detrimental to classification of the target language. We analyze this further
in Section \ref{acl:analysism}. Note that
in this configuration, we do not need to update the original vector
spaces, which would be problematic with such small training data.

The projection quality is ensured by minimizing the mean squared
error\footnote{We omit parameters in equations for better
  readability.}\footnote{We also experimented with cosine distance,
  but found that it performed worse than Euclidean distance.}
\begin{equation}
\textrm{MSE} = \dfrac{1}{n} \sum_{i=1}^{n} (\mathbf{z_{i}} - \mathbf{\hat{z}_{i}})^{2}\,,
\end{equation}
where $\mathbf{z_{i}} = S_{s_{i}} \cdot M$ is the dot product of the embedding for source word $s_{i}$ and the source projection matrix and $\mathbf{\hat{z}_{i}} = T_{t_{i}} \cdot M'$ is the same for the target word $t_{i}$.

\subsubsection{Sentiment Classification}
\label{acl:sentiment}

We add a second training objective to optimize the projected source
vectors to predict the sentiment of source phrases. This inevitably
changes the projection characteristics of the matrix $M$, and
consequently $M'$ and encourages $M'$ to learn to predict sentiment
without any training examples in the target language.

In order to train $M$ to predict sentiment, we require a source-language
corpus  $\Csource = \{(x_{1}, y_{1}),
(x_{2}, y_{2}), \ldots, (x_{i}, y_{i})\}$ where each sentence $x_{i}$ 
is associated with a label $y_{i}$.

For classification, we use a two-layer feed-forward averaging network,
loosely following \shortcite{Iyyer2015} 
\footnote{Our model employs a
  linear transformation after the averaging layer instead of including
  a non-linearity function. We choose this architecture because the
  weights $M$ and $M'$ are also used to learn a linear cross-lingual
  projection.}. For a sentence $x_{i}$ we take the word embeddings
from the source embedding $S$ and average them to $\mathbf{a}_{i} \in
\R^{d}$. We then project this vector to the joint bilingual space
$\mathbf{z}_{i} = \mathbf{a}_{i} \cdot M$. Finally, we pass
$\mathbf{z}_{i}$ through a softmax layer $P$ to obtain the prediction
$\hat{y}_{i} = \textrm{softmax} ( \mathbf{z}_{i} \cdot P)$.

To train our model to predict sentiment, we minimize the cross-entropy
error of the predictions\\[-\baselineskip]
\begin{equation} 
H = - \sum_{i=1}^{n} y_{i} \log \hat{y_{i}} - (1 - y_{i}) \log (1 - \hat{y_{i}})\,.
\end{equation}

\subsubsection{Joint Learning}
\label{acl:joint}
In order to jointly train both the projection component and the
sentiment component, we combine the two loss functions to optimize the
parameter matrices $M$, $M'$, and $P$ by
\begin{equation}
J =\kern-5mm
\sum_{(x,y) \in \Csource}\kern1mm \sum_{(s,t) \in L}\kern0mm \alpha H(x,y)
+ (1 - \alpha) \cdot \textrm{MSE}(s,t)\,,
\end{equation}
where $\alpha$ is a hyperparameter that weights sentiment loss vs.\ projection loss.

\subsubsection{Target-language Classification}
For inference, we classify sentences from a target-language corpus
$\Ctarget$.
As in the training procedure, for each sentence, we take the word
embeddings from the target embeddings $T$ and average them to
$\mathbf{a}_{i} \in \R^{d}$. We then project this vector to the joint
bilingual space $\mathbf{\hat{z}}_{i} = \mathbf{a}_{i} \cdot
M'$. Finally, we pass $\mathbf{\hat{z}}_{i}$ through a softmax layer
$P$ to obtain the prediction $\hat{y}_{i} = \textrm{softmax} (
\mathbf{\hat{z}}_{i} \cdot P)$.

\subsection{Targeted Model}
\label{targetedprojection}

\begin{figure}
  \centering
  \includegraphics[width=.8\textwidth]{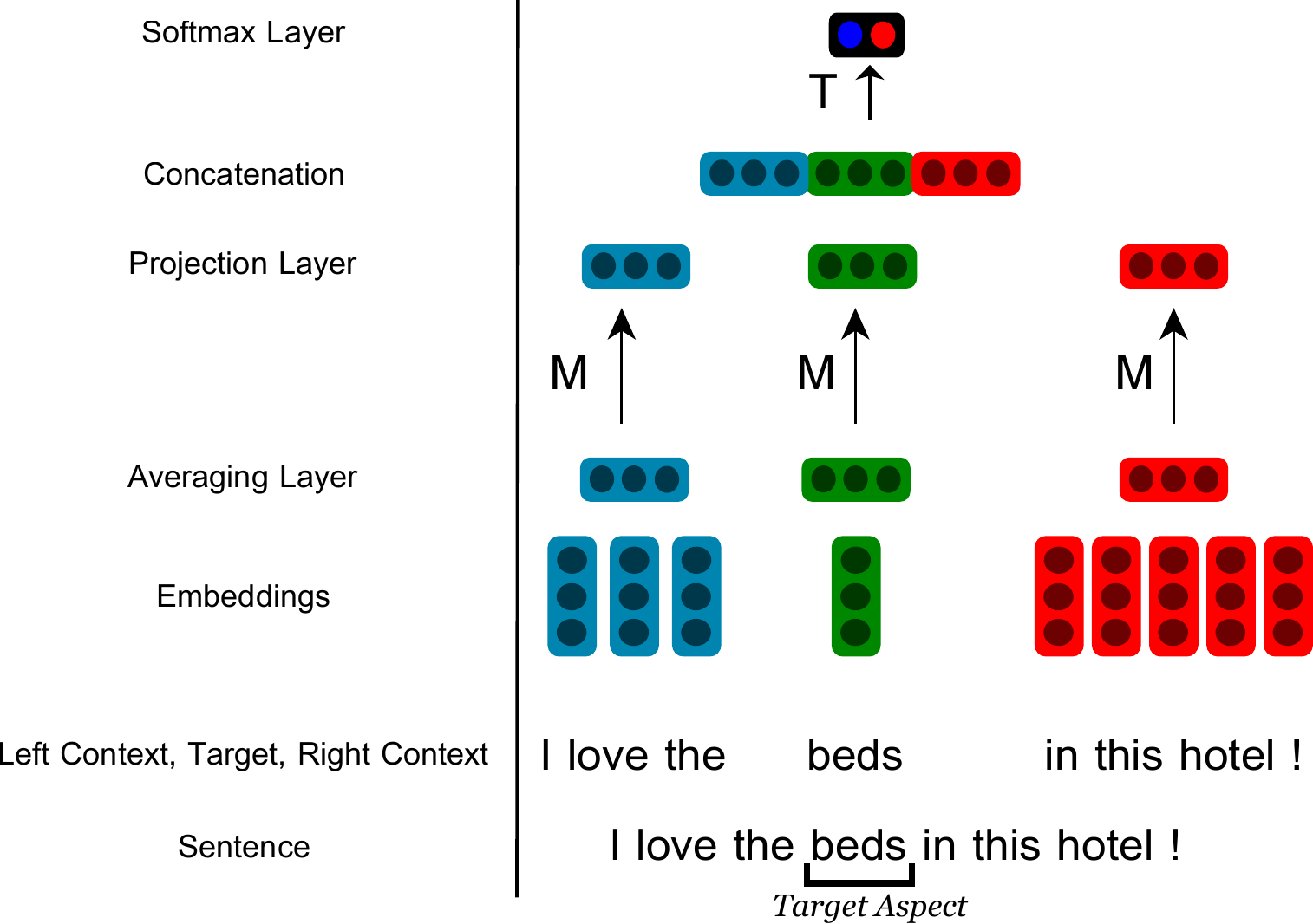}
  \caption{The \splitsent adaptation of our \blse model to targeted sentiment analysis. At test time, we replace the matrix $M$ with the matrix $M^{'}$.}
  \label{beyond:fig:model}
\end{figure}

In our targeted model, we assume that the list of sentiment targets as they
occur in the text is given. These can be extracted previously either
by using domain knowledge \shortcite{Liu2005}, by using a named entity
recognizer \shortcite{Zhang2015} or by using a number of aspect extraction
techniques \shortcite{Zhou2012}. Given these targets, the task is reduced
to classification. However, what remains is how to represent the
target, to learn to subselect the information from the context which
is relevant, how to represent this contextual information, and how to
combine these representations in a meaningful way that enables us to
classify the target reliably.

Our approach to adapt the \blse model to targeted sentiment analysis,
which we call \splitsent (depicted in Figure \ref{beyond:fig:model}),
is similar to the method proposed by \shortciteA{Zhang2016} for gated
recurrent networks. For a sentence with a target $a$, we split the
sentence at $a$ in order to get a left and right context, $\llcon(a)$
and $\rcon(a)$ respectively.

Unlike the approach from \shortciteA{Zhang2016}, we do not use
recurrent neural networks to create a feature vector, as
\shortciteA{Atrio2019} showed that, in cross-lingual
setups, they overfit too much to word order and source-language
specific information to perform well on our tasks. Therefore, we
instead average each left context $\llcon(a_i)$, right context
$\rcon(a_i)$, and target $a_{i}$ separately. Although averaging is a
simplified approach to create a compositional representation of a
phrase, it has been shown to work well for sentiment
\shortcite{Iyyer2015,Barnes2017}. After creating a single averaged
vector for the left context, right context, and target, we concatenate
them and use these as input for the softmax classification layer $T \in
\R^{d \times 3}$, where $d$ is the dimensionality of the input vectors. The model is trained on the source language sentiment data using $M$ to project, and then tested by replacing $M$ with $M^{'}$, similar to the sentence-level model.

\section{Experiments}
\label{sec:experiments}

In this section, we describe the resources and datasets, as well as
the experimental setups used in both the sentence-level (Experiment 1
in Subsection~\ref{experiment1}) and targeted (Experiment 2 in
Subsection~\ref{experiment2}) experiments.

\subsection{Datasets and Resources}

The number of datasets and resources for under-resourced languages are
limited. Therefore, we choose a mixture of resource-rich and under-resourced
languages for our experiments. We treat the resource-rich languages as if
they were under-resourced by using similar amounts of parallel data.

\subsubsection{Sentence-level Datasets}

\begin{table}
\centering
\begin{tabular}{lrrrrr}
\toprule
    & & \multicolumn{1}{c}{EN} & \multicolumn{1}{c}{ES} & \multicolumn{1}{c}{CA} & \multicolumn{1}{c}{EU}\\
\cmidrule(rl){2-2}\cmidrule(l){3-3}\cmidrule(l){4-4}\cmidrule(l){5-5}\cmidrule(l){6-6}
 \multirow{3}{*}{\rt{Binary}}
 &$+$   & 1258 & 1216 & 718  & 956    \\
 &$-$   & 473 & 256 & 467  & 173   \\
 &\textit{Total}    &1731 & 1472  &   1185   & 1129        \\
\cmidrule(rl){2-2}\cmidrule(l){3-3}\cmidrule(l){4-4}\cmidrule(l){5-5}\cmidrule(l){6-6}
 \multirow{5}{*}{\rt{4-class}}
 &$++$   & 379 & 370  & 256  & 384 \\
 &$+$    & 879 & 846  & 462   & 572 \\
 &$-$    & 399 & 218  & 409   & 153 \\
 &$--$   &  74 & 38   & 58    & 20  \\
 &\textit{Total}     & 1731  & 1472     & 1185      &   1129  \\
\bottomrule
\end{tabular}
\caption{Statistics for the OpeNER English (EN) and Spanish (ES) 
as well as the MultiBooked Catalan (CA) and Basque (EU) datasets.}
\label{blse:datasetstats}
\end{table}

To evaluate our proposed model at sentence-level, we conduct experiments using four
benchmark datasets and three bilingual combinations. We use the OpeNER
English and Spanish datasets \shortcite{Agerri2013} and the MultiBooked
Catalan and Basque datasets \cite{Barnes2018a}. All
datasets contain hotel reviews which are annotated for targeted
sentiment analysis. The labels include \textit{Strong Negative}
($--$), \textit{Negative} ($-$), \textit{Positive} ($+$), and
\textit{Strong Positive} ($++$). We map the aspect-level annotations
to sentence level by taking the most common label and remove instances
of mixed polarity. We also create a binary setup by combining the
strong and weak classes. This gives us a total of six experiments. The
details of the sentence-level datasets are summarized in
Table~\ref{blse:datasetstats}.
\begin{table}
\centering
\begin{tabular}{lrrr}
\toprule
   & \multicolumn{1}{c}{Spanish} & \multicolumn{1}{c}{Catalan} & \multicolumn{1}{c}{Basque}\\
\cmidrule(r){1-1}\cmidrule(rl){2-2}\cmidrule(l){3-3}\cmidrule(l){4-4}
 Sentences & 23 M  &    9.6 M   &   0.7 M  \\
 Tokens  & 610 M & 183 M  & 25 M \\
 Embeddings & 0.83 M &  0.4 M & 0.14 M \\
\bottomrule
\end{tabular}
\caption{Statistics for the Wikipedia corpora and monolingual vector spaces.}
\label{stats:wikis}
\end{table}
For each of the experiments, we take 70 percent of the data for
training, 20 percent for testing and the remaining 10 percent are used
as development data for tuning meta-parameters.

\subsubsection{Targeted Datasets}
\label{targeteddatasets}

We use the following corpora to set up the experiments in which we
train on a source language corpus $C_{S}$ and test on a target
language corpus $C_{T}$. Statistics for all of the corpora are shown
in Table~\ref{stats:datasets}. We include a binary
classification setup, where neutral has been removed and strong
positive and strong negative have been mapped to positive and
negative, as well as a multiclass setup, where the original
labels are used.

\textbf{OpeNER Corpora:} The OpeNER corpora \shortcite{Agerri2013} are
composed of hotel reviews, annotated for aspect-based sentiment. Each
aspect is annotated with a sentiment label (Strong Positive, Positive,
Negative, Strong Negative). We perform experiments with the English
and Spanish versions.

\textbf{MultiBooked Corpora:} The MultiBooked corpora
\shortcite{Barnes2018a} are also hotel reviews annotated in the same way as
the OpeNER corpora, but in Basque and Catalan. These
corpora allow us to observe how well each approach performs on
low-resource languages.

\textbf{SemEval 2016 Task 5:} We take the English and Spanish
restaurant review corpora made available by the organizers of the
SemEval event \shortcite{Pontiki2016}. These corpora are annotated for
three levels of sentiment (positive, neutral, negative).

\textbf{USAGE Corpora:} The USAGE corpora \shortcite{Klinger2014a} are
Amazon reviews taken from a number of different items, and are
available in English and German. Each aspect is annotated for three
levels of sentiment (positive, neutral, negative). As the corpus has
two sets of annotations available, we take the 
annotations from annotator 1 as the gold standard.

\begin{table}
\centering\small
\newcommand{\sepp}{\cmidrule(r){2-2}\cmidrule(lr){3-3}\cmidrule(lr){4-4}\cmidrule(lr){6-6}\cmidrule(lr){7-7}\cmidrule(lr){8-8}\cmidrule(lr){9-9}\cmidrule(l){10-10}}
\setlength\tabcolsep{3.1pt}
\begin{tabular}{llrrp{2mm}rrrrr}
\toprule
& & \multicolumn{2}{c}{Binary} && \multicolumn{5}{c}{Multiclass}\\
\cmidrule(lr){3-4}\cmidrule(lr){6-10}
& & \ccc{$+$} & \ccc{$-$} && \ccc{$++$} & \ccc{$+$} & \ccc{0} & \ccc{$-$} & \ccc{$--$} \\
\sepp
\multirow{2}{*}{OpeNER} & EN & 1658 & 661 && 472 & 1132 &&556 & 105\\
& ES & 2404 & 446 && 813 & 1591 & & 387 & 59\\
\sepp
\multirow{2}{*}{MultiBooked} & CA & 1453 & 883 && 645 & 808 & & 741 & 142\\
& EU & 1461 & 314 && 686 & 775 & & 273 & 41 \\
\sepp
\multirow{2}{*}{SemEval} & EN & 2268 & 953 && & 2268 & 145 & 953 \\
& ES & 2675 & 948 && & 2675 & 168 & 948\\
\sepp
\multirow{2}{*}{USAGE} & EN & 2985 & 1456 && & 2985 & 34 & 1456\\
& DE & 3115 & 870 &&& 3115 & 99 & 870\\
\bottomrule
\end{tabular}
\caption{Number of aspect-polarity tuples for the targeted datasets.}
\label{stats:datasets}
\end{table}

\subsubsection{Resources}

\paragraph{Monolingual Word Embeddings}
\label{embeddings}
For \blse, \vecmap, \muse, and \mt, we require monolingual vector
spaces for each of our languages. For English, we use the publicly
available GoogleNews
vectors\footnote{\label{word2vecfootnote}\url{https://code.google.com/archive/p/word2vec/}}. For
Spanish, Catalan, and Basque, we train skip-gram embeddings using the
Word2Vec toolkit\footnote{\url{https://code.google.com/archive/p/word2vec/}} with 300 dimensions, subsampling
of $10^{-4}$, window of 5, negative sampling of~15 based on a 2016
Wikipedia
corpus\footnote{\url{http://attardi.github.io/wikiextractor/}}
(sentence-split, tokenized with IXA pipes \shortcite{Agerri2014} and
lowercased). The statistics of the Wikipedia corpora are given in
Table \ref{stats:wikis}.

\paragraph{Bilingual Lexicon}
\label{transdict}
For \blse, \vecmap, \muse, and \barista, we also require a bilingual
lexicon. We use the sentiment lexicon from \shortciteA{HuandLiu2004} (to
which we refer in the following as Hu and Liu) and its translation into
each target language. We translate the lexicon using Google Translate
and exclude multi-word expressions.\footnote{Note that we only do that
  for convenience. Using a machine translation service to generate
  this list could easily be replaced by a manual translation, as the
  lexicon is comparably small.} This leaves a dictionary of 5700
translations in Spanish, 5271 in Catalan, and 4577 in Basque. We set
aside ten percent of the translation pairs as a development set in
order to check that the distances between translation pairs not seen
during training are also minimized during training.

\subsection{Setting for Experiment 1: Sentence-level Classification}
\label{experiment1}
We compare \blse (Sections \ref{acl:crosslingual}--\ref{acl:joint}) to
\vecmap, \muse, and \barista
(Section~\ref{sec:previouswork}) as baselines, which have similar data
requirements and to machine translation (\mt) and monolingual (\mono)
upper bounds which request more resources.  For all models (\mono,
\mt, \vecmap, \muse, \barista), we take the average of the word embeddings
in the source-language training examples and train a linear
SVM\footnote{LinearSVC implementation from scikit-learn.}. We report
this instead of using the same feed-forward network as in \blse as it
is the stronger upper bound. We choose the parameter $c$ on the target
language development set and evaluate on the target language test set.

\textbf{Upper Bound \mono.} We set an empirical upper bound by
training and testing a linear SVM on the target language data. Specifically,
we train the model on
the averaged embeddings from target language training data, tuning the
$c$ parameter on the development data. We test on the target language
test data.

\textbf{Upper Bound \mt.}
To test the effectiveness of machine translation, we translate all of
the sentiment corpora from the target language to English using the
Google Translate
API\footnote{\url{https://translate.google.com}}. Note that this
approach is not considered a baseline, as we assume not to have
access to high-quality machine translation for low-resource languages
of interest.

\textbf{Baseline \unsup}
We compare with the unsupervised statistical machine translation approach proposed by \shortciteA{artetxe2018emnlp}. This approach uses a self-supervised method to create bilingual phrase embeddings which then populates a phrase table. Monolingual n-gram language models and an unsupervised variant of MERT are used to create a MT model which is improved through iterative backtranslation. We use the Wikipedia corpora from Section \ref{embeddings} to create the unsupervised SMT system between English and the target languages and run the training proceedure with default parameters. Finally, we translate all test examples in the target languages to English.

\textbf{Baseline \vecmap.}
We compare with the approach proposed by \shortciteA{Artetxe2016} which
has shown promise on other tasks, \eg, word similarity. In order to
learn the projection matrix $W$, we need translation pairs. We use the
same word-to-word bilingual lexicon mentioned in Section
\ref{acl:crosslingual}. We then map the source vector space $S$ to the
bilingual space $\hat{S} = SW$ and use these embeddings.

\textbf{Baseline \muse.}  This baseline is similar to \vecmap but incorporates
and adversarial objective as well as a localized scaling objective, which 
further improve the orthogonal refinement so that the two language
spaces are even more similar.

\textbf{Baseline \barista.}  The approach proposed by
\shortciteA{Gouws2015taskspecific} is another appropriate baseline, as 
it fulfills the same data requirements as the projection methods. 
The bilingual lexicon used to
create the pseudo-bilingual corpus is the same word-to-word bilingual
lexicon mentioned in Section \ref{acl:crosslingual}. We follow the
authors' setup to create the pseudo-bilingual corpus. We create
bilingual embeddings by training skip-gram embeddings using the
Word2Vec toolkit on the pseudo-bilingual corpus using the same
parameters from Section \ref{embeddings}.

\textbf{Our method: BLSE.}
Our model, \blse, is implemented in Pytorch \shortcite{Pytorch} and
the word embeddings are initialized with the pretrained word embeddings $S$ and $T$
mentioned in Section \ref{embeddings}. We use the word-to-word
bilingual lexicon from Section \ref{transdict}, tune the
hyperparameters $\alpha$, training epochs, and batch size on the
target development set and use the best hyperparameters achieved on
the development set for testing. ADAM \shortcite{Kingma2014a} is used in
order to minimize the average loss of the training batches.

\textbf{Ensembles.} In order to evaluate to what extent each projection
model adds complementary information to the machine translation
approach, we create an ensemble of \mt and each projection
method (\blse, \vecmap, \muse, \barista).  A random forest
classifier is trained on the predictions from \mt and each of these
approaches.

\subsection{Setting for Experiment 2: Targeted Classification}
\label{experiment2}

For the targeted classification experiment, we compare the same models
mentioned above, but adapted to the setting using the \splitsent method 
from Section \ref{targetedprojection}.

A simple majority baseline sets the lower bound, while the \mt-based 
model serves as an  upper
bound. We assume our models to perform between these two, as we do not
have access to the millions of parallel sentences required to perform
high-quality \mt and particularly aim at proposing a method which is
less resource-hungry.

\paragraph{Simplified Models: Target only and Context only}
We hypothesize that cross-lingual approaches are particularly
error-prone when evaluative phrases and words are wrongly
predicted. In such settings, it might be beneficial for a model to put
emphasis on the target word itself and learn a prior
distribution of
sentiment for each target independent of the context. For example, if you 
assume that all mentions of Steven Segal are negative in movie reviews,
it is possible to achieve good results \shortcite{Bird2009}. On the other hand,
it may be that there are not enough examples of target-context pairs, and
that it is better to ignore the target and concentrate only on the contexts.

To analyze
this, we compare our model to two simplified versions. In addition,
this approach enables us to gain insight in the source of relevant
information. The first is \asponly, which means that we use the model in
the same way as before but ignore the context completely. This serves
as a tool to understand how much model performance originates from the
target itself.

In the same spirit, we use a \cononly model, which ignores
the target by constraining the parameters of all target phrase embeddings
to be the same. This approach might be beneficial over our initial
model if the prior distribution between targets was similar and the
context actually carries the relevant information.

\paragraph{Baseline: Sentence Assumption}
As the baseline for each projection method, we assume all targets in
each sentence respectively to be of the same polarity (\sent). This is
generally an erroneous assumption, but can give good results if all of
the targets in a sentence have the same polarity. In addition, this
baseline provides us with the information about whether the models are
able to handle information from different positions in the text.

\section{Results}
\label{sec:results}

\subsection{Experiment 1: Sentence-level Classification}

\begin{table*}
\definecolor{blue}{cmyk}{0.2,0,0,0.1}
\renewcommand{\hl}[1]{{\textcolor{violet}{\emph{#1}}}}
\newcommand{\best}[1]{\textbf{\setlength{\fboxsep}{1pt}\fbox{#1}}}
\newcommand{\sep}{\cmidrule(r){3-3}\cmidrule(r){4-4}\cmidrule(r){5-5}\cmidrule(r){6-6}\cmidrule(r){7-7}\cmidrule(r){8-8}\cmidrule(r){9-9}\cmidrule(r){10-10}\cmidrule(r){11-11}\cmidrule(r){12-12}}
\newcommand{\sepd}{\cmidrule(lr){4-6}\cmidrule(l){7-9}}
 \centering
 \renewcommand*{\arraystretch}{1.0}
\setlength\tabcolsep{0.5mm}
 \begin{tabular}{llccccccccccc}
 \toprule
 &&\multicolumn{2}{c}{Upper Bounds}& \multicolumn{1}{c}{}&\multicolumn{3}{c}{Baselines}&\multicolumn{4}{c}{Ensemble}\\
\cmidrule(r){3-4}\cmidrule(r){6-9}\cmidrule(r){10-13}
&&\mono & \mt & \small{\textbf{BLSE}}  & \unsup & \vecmap & \muse & \barista & \vecmap & \muse & \barista & \blse \\
\sep

\multirow{3}{*}{\rt{Binary}}  & ES & 73.5 & 79.0 & \hl{*74.6} & 76.8  & 67.1 & 73.4 & 61.2 & 62.6 &  58.7   & 56.0 & \textbf{80.3} \\
                              & CA & 79.2 & 77.2  & \hl{*72.9} & 79.4 & 60.7 & 71.1 &60.1 & 63.3 &  64.3   & 62.5 & \textbf{85.0}\\
                              & EU & 69.8 & 69.4 &  \hl{*69.3} & 65.5 & 45.6 & 59.8 &54.4 & 66.4 &  68.4   & 49.8 & \textbf{73.5}\\
\sep

\multirow{3}{*}{\rt{4-class}}  & ES & 45.5 & 48.8 & \hl{41.2} & 49.1 & 34.9 &  37.1   &39.5 & 43.8 &   49.3   & 47.1 & \textbf{50.3}\\
                              & CA & 49.9 & 52.7  &35.9 & 47.7 & 23.0 & 39.0 & 36.2 & 47.6 &      52.0 & 53.0 & \textbf{53.9}\\
                              & EU & 47.1 & 43.6 & 30.0 & 39.3  & 21.3 & 25.8 & 33.8 & 49.9 &      46.4 & 47.8 & \textbf{50.5}\\

 \bottomrule
 \end{tabular}
 \caption{Macro \F of four models
 trained on English and tested on Spanish (ES), Catalan (CA), and
 Basque (EU). The \textbf{bold} numbers show the best results for
 each metric per column and the \hl{highlighted} numbers show where
 \blse is better than the other projection methods, \vecmap, \muse, and
 \barista (* p $<$ 0.01).}
 \label{results}
\end{table*}

In Table \ref{results}, we report the results of all four
methods. Our method outperforms the other projection methods (the
baselines \vecmap, \muse, and \barista) on four of the six experiments
substantially. It performs only slightly worse than the more
resource-costly upper bounds (\mt and \mono).
This is especially noticeable for the binary
classification task, where \blse performs nearly as well as machine
translation and significantly better than the other methods. \unsup also performs similarly to \blse on the binary tasks, while giving stronger performance on the 4-class setup. We
perform approximate randomization tests \shortcite{Yeh2000} with 10,000
runs and highlight the results that are statistically significant (*p
$<$ 0.01) in Table \ref{results}.

In more detail, we see that \mt generally performs better than the
projection methods (79--69 \F on binary, 52--44 on 4-class). \blse
(75--69 on binary, 41--30 on 4-class) has the best performance of the
projection methods and is comparable with \mt on the binary setup,
with no significant difference on binary Basque. \vecmap (67--46 on
binary, 35--21 on 4-class) and \barista (61--55 on binary, 40--34 on
4-class) are significantly worse than \blse on all experiments except
Catalan and Basque 4-class. \muse (67--62 on binary, 45--34 on 4-class)
performs better than \vecmap and \barista. On the binary experiment, \vecmap
outperforms \barista on Spanish (67.1 vs.\ 61.2) and Catalan (60.7
vs.\ 60.1) but suffers more than the other methods on the four-class
experiments, with a maximum \F of 34.9. \barista is relatively stable
across languages. \unsup performs well across experiments (76--65 on binary, 49--39 on 4-class), even performing better than \mt on both Catalan tasks and Spanish 4-class.

The \ensemble of \mt and \blse performs the best, which shows that \blse
adds complementary information to \mt. Finally, we note that all
systems perform worse on Basque. This is
presumably due to the increased morphological complexity of Basque,
as well as its lack of similarity to the source language English (Section \ref{deployment:langsimilarity}).

\subsubsection{Model and Error Analysis}

We analyze three aspects of our model in further detail: 1) where most mistakes originate, 2) the effect of the bilingual lexicon, and 3) the effect and necessity of the target-language projection matrix $M'$. 

\subsubsection{Phenomena}

In order to analyze where each model struggles,
we categorize the mistakes and annotate all of the test
phrases with one of the following error classes: vocabulary (voc),
adverbial modifiers (mod), negation (neg), external knowledge (know) or other. Table \ref{acl:erroranalysis}
shows the results.

\textbf{Vocabulary:} The most common way to express sentiment in hotel
reviews is through the use of polar adjectives (as in ``the room was
great'') or the mention of certain nouns that are desirable (``it had a
pool''). Although this phenomenon has the largest total number of
mistakes (an average of 72 per model on binary and 172 on 4-class), it
is mainly due to its prevalence. \mt performed the best on the test examples which
according to the annotation require a correct understanding of the vocabulary
 (81 \F on binary /54 \F on 4-class),
with \blse (79/48) slightly worse. \muse (76/23), \vecmap (70/35), and \barista
(67/41) perform worse. This suggests that \blse is better than \muse, \vecmap and
\barista at transferring sentiment of the most important sentiment
bearing words.

\textbf{Negation:} Negation is a well-studied phenomenon in sentiment
analysis \shortcite{Pang2002,Wiegand2010,Zhu2014,Reitan2015}
. Therefore, we
are interested in how these four models perform on phrases that
include the negation of a key element, for example ``In general, this
hotel isn't bad". We would like our models to recognize that the
combination of two negative elements ``isn't" and ``bad" lead to a
\textit{Positive} label.

\begin{table}[]
\newcommand{\sep}{\cmidrule(r){1-1}\cmidrule(rl){2-2}\cmidrule(lr){3-3}\cmidrule(lr){4-4}\cmidrule(rl){5-5}\cmidrule(rl){6-6}\cmidrule(rl){7-7}\cmidrule(l){8-8}}
\centering
\renewcommand*{\arraystretch}{0.8}
\setlength\tabcolsep{2.0mm}
\begin{tabular}{llrrrrrr}
\toprule
Model & & \multicolumn{1}{c}{\rt{voc}} & \multicolumn{1}{c}{\rt{mod}} & \multicolumn{1}{c}{\rt{neg}} & \multicolumn{1}{c}{\rt{know}}&\multicolumn{1}{c}{\rt{other}}&\multicolumn{1}{c}{\rt{\textit{total}}}\\
\sep\multirow{2}{*}{\footnotesize{\mt}}
 &bi & 49 & 26 & 19 & 14 & 5 & \textbf{113} \\
 &4 & 147 & 94 & 19 & 21 & 12 & \textbf{293} \\
 \sep\multirow{2}{*}{\footnotesize{\unsup}}
 &bi & 65 & 31 & 21 & 17 & 7 & \textbf{141} \\
 &4 & 170 & 120 & 27 & 26 & 15 & \textbf{358} \\
 \sep\multirow{2}{*}{\footnotesize{\muse}}
 &bi & 75 & 38 & 17 & 18 & 8 & \textbf{156} \\
 &4 & 195  & 137 & 27  & 22 & 28 & \textbf{409} \\
\sep \multirow{2}{*}{\footnotesize{\vecmap}}
 &bi & 80 &44 &27 &14 &7 &\textbf{172}\\
 &4 & 182 &141 &19 &24 &19 &\textbf{385}\\
\sep \multirow{2}{*}{\footnotesize{\barista}}
 &bi & 89 &41 &27 &20 &7 &\textbf{184} \\ 
 &4 & 191 &109 &24 &31 &15 &\textbf{370} \\
\sep \multirow{2}{*}{\footnotesize{\blse}}
 &bi & 67 &45 &21 &15 &8 &\textbf{156}\\
 &4 & 146 &125 &29 &22 &19 &\textbf{341}\\
\bottomrule
\end{tabular}
\caption[\blse error analysis]{Error analysis for different phenomena for the binary (bi) and multi-class (4) setups. See text for
  explanation of error classes.}
\label{acl:erroranalysis}
\end{table}

Given the simple classification strategy,
all models perform relatively well on phrases with negation (all reach nearly
60 \F in the binary setting). However, while
\blse performs the best on negation in the binary setting (82.9
\F), it has more problems with negation in the 4-class setting
(36.9 \F).

\textbf{Adverbial Modifiers:} Phrases that are modified by an adverb,
\eg, the food was \textit{incredibly} good, are important for the
four-class setup, as they often differentiate between the base and
\textit{Strong} labels. In the binary case, all models reach more than
55 \F. In the 4-class setup, \blse only achieves 27.2 \F compared to
46.6 or 31.3 of \mt and \barista, respectively. Therefore, presumably,
our model does currently not capture the semantics of the target
adverbs well. This is likely due to the fact that it assigns too much
sentiment to functional words (see Figure \ref{acl:fig:tsne}). \muse performs
poorly on modified examples (20.3 \F).

\textbf{External Knowledge Required:} These errors are difficult for
any of the models to get correct. Many of these include numbers which
imply positive or negative sentiment (350 meters from the beach is
\textit{Positive} while 3 kilometers from the beach is
\textit{Negative}). \blse performs the best (63.5 \F) while \mt
performs comparably well (62.5). \barista performs the worst (43.6).

\textbf{Binary vs. 4-class:} All of the models suffer when moving from
the binary to 4-class setting; an average of 26.8 in macro \F for \mt,
31.4 for \vecmap, 22.2 for \barista, 34.1 for \muse, and 36.6 for \blse. The
vector projection methods (\vecmap, \muse, and \blse) suffer the most,
suggesting that they are currently more apt for the binary setting.

\subsubsection{Effect of Bilingual Lexicon}

We analyze how the number of translation pairs affects our model. We train on the 4-class Spanish setup using the best hyper-parameters from the previous experiment.

\begin{figure}[]
\begin{center}
\includegraphics[width = 4in]{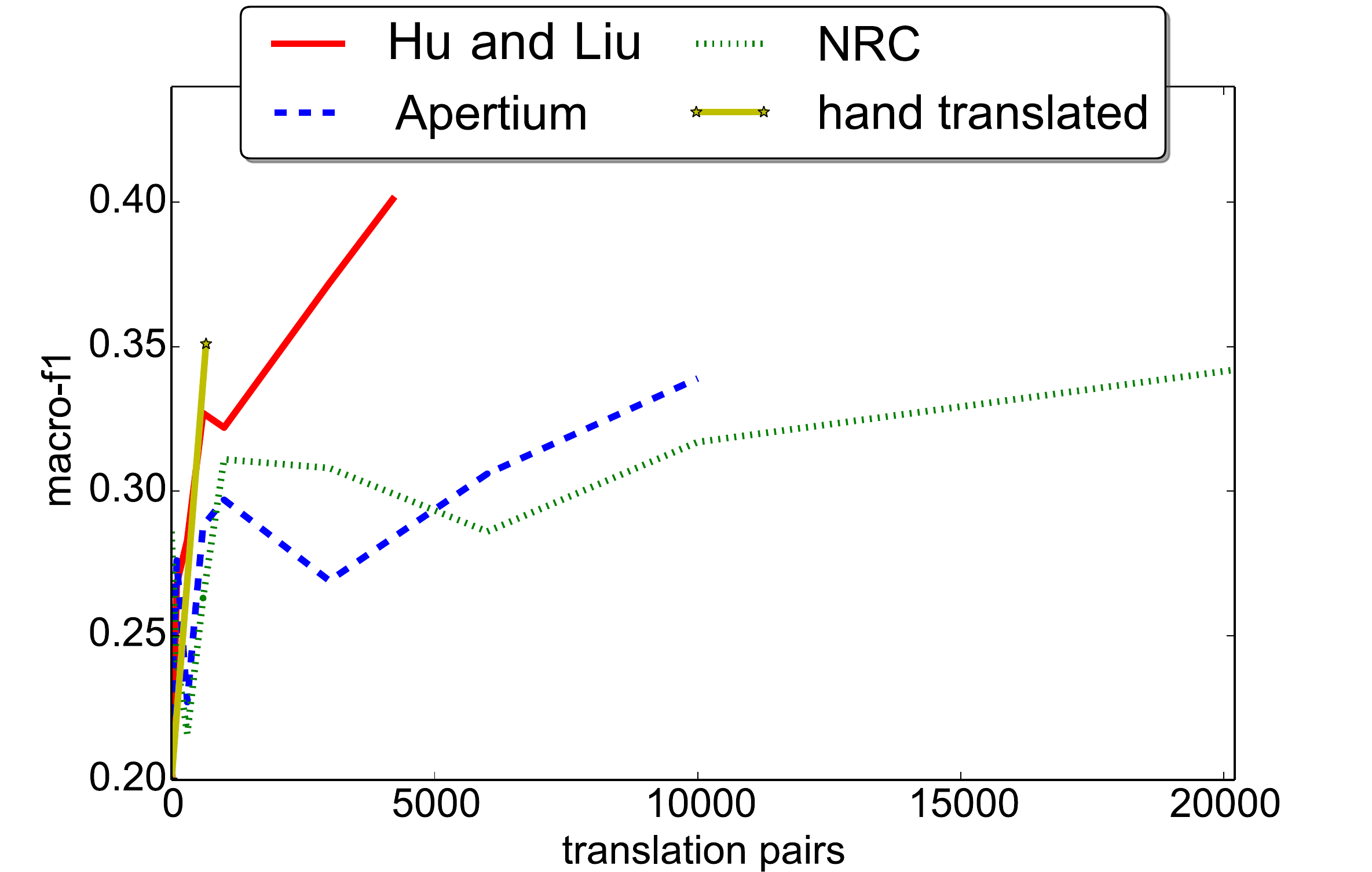}
\caption[Analysis of projection lexicon]{Macro \F for translation pairs in the Spanish 4-class setup. Training with the expanded hand translated lexicon and machine-translated Hu and Liu lexicon gives a macro \F that grows constantly with the number of translation pairs. Despite having several times more training data, the Apertium and NRC translation dictionaries do not perform as well. 
}
\label{acl:fig:transdict}
\end{center}
\end{figure}

Research into projection techniques for bilingual word embeddings
\shortcite{Mikolov2013translation,Lazaridou2015,Artetxe2016} often uses a
lexicon of the most frequent 8--10 thousand words in English and their
translations as training data. We test this approach by taking the
10,000 word-to-word translations from the Apertium English-to-Spanish
dictionary\footnote{\url{http://www.meta-share.org}}. We also use the
Google Translate API to translate the NRC hashtag sentiment lexicon
\shortcite{Mohammad2013} and keep the 22,984 word-to-word
translations. We perform the same experiment as above and vary the
amount of training data from 0, 100, 300, 600, 1000, 3000, 6000,
10,000 up to 20,000 training pairs. Finally, we compile a small hand
translated dictionary of 200 pairs, which we then expand using target
language morphological information, finally giving us 657 translation
pairs\footnote{The translation took approximately one hour. We can
  extrapolate that manually translating a sentiment lexicon the size of
  the Hu and Liu lexicon would take no more than 5 hours. }.  The macro
\F score for the Hu and Liu dictionary climbs constantly with the
increasing translation pairs. Both the Apertium and NRC dictionaries
perform worse than the translated lexicon by Hu and Liu, while the
expanded hand translated dictionary is competitive, as shown in
Figure \ref{acl:fig:transdict}.

While for some tasks, \eg, bilingual lexicon induction, using the most
frequent words as translation pairs is an effective approach, for
sentiment analysis, this does not seem to help. Using a translated
sentiment lexicon, even if it is small, gives better results.

\subsubsection{Analysis of $M'$}
\label{acl:analysism}

The main motivation for using two projection matrices $M$ and $M'$ is
to allow the original embeddings to remain stable, while the projection
matrices have the flexibility to align translations and separate these
into distinct sentiment subspaces. To justify this design decision empirically, we perform
an experiment to evaluate the actual need for the target language
projection matrix $M'$: We create a simplified version of our model
without $M'$, using $M$ to project from the source to target and then
$P$ to classify sentiment.

\begin{figure}[]
\centering
\includegraphics[width=0.5\textwidth]{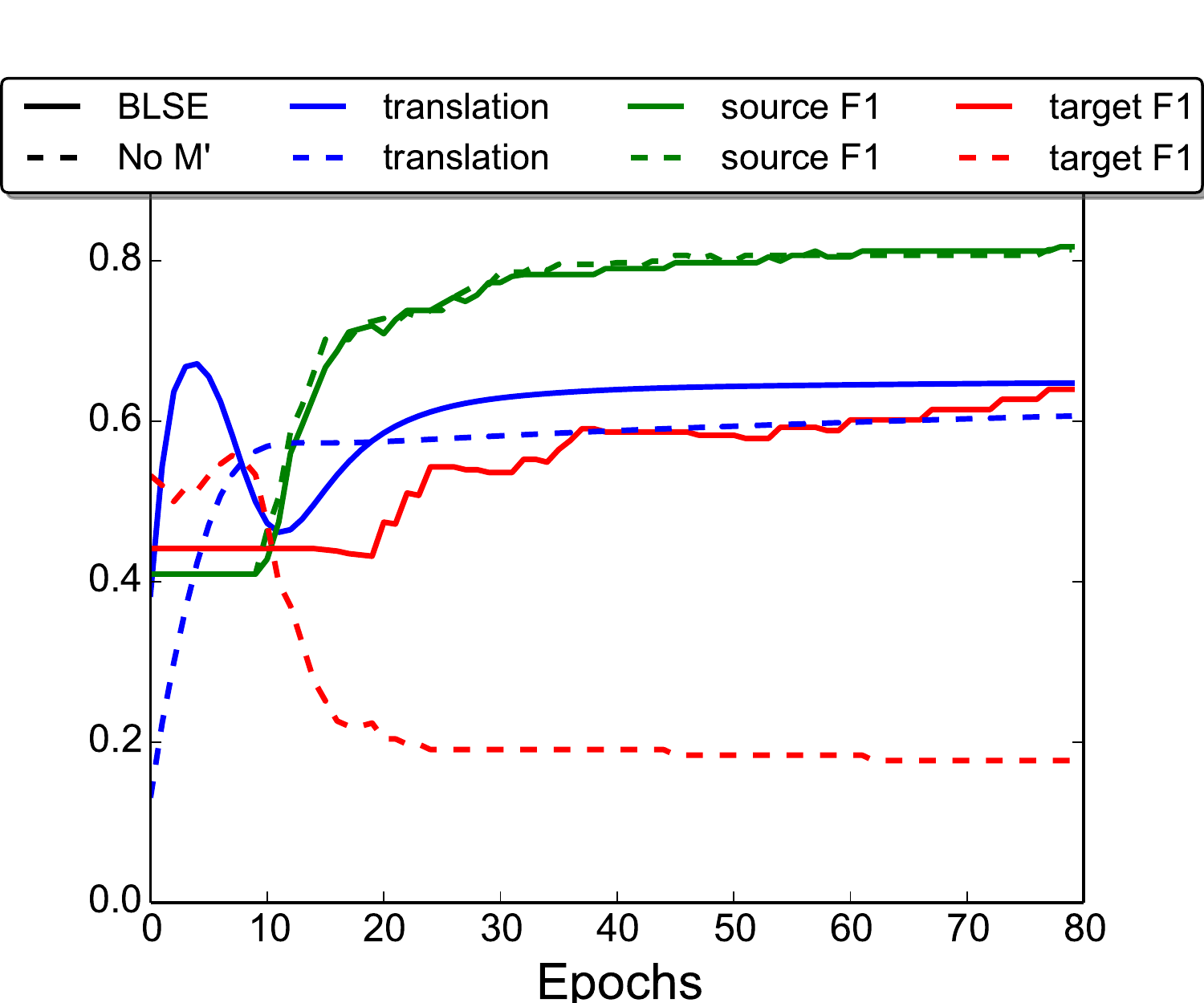}
\caption[Analysis of $M'$]{\blse model (solid lines) compared to a variant without
  target language projection matrix $M'$ (dashed lines).
  ``Translation'' lines show the average cosine similarity between
  translation pairs. The remaining lines show \F scores for the source
  and target language with both variants of \blse. The modified model
  cannot learn to predict sentiment in the target language (red
  lines).  This illustrates the need for the second projection matrix
  $M'$.}
\label{acl:fig:nomprime}
\end{figure}

The results of this model are shown in
Figure~\ref{acl:fig:nomprime}. The modified model does learn to predict
in the source language, but not in the target language. This confirms
that $M'$ is necessary to transfer sentiment in our model.

\subsubsection{No projection}

Additionally, we provide an analysis of a similar model to ours, but which uses $M = \R^{d, o}$ and $M' = \R^{d', o}$, where $d$ ($d'$) is the dimensionality of the original embeddings and $o$ is the label size, to directly model crosslingual sentiment, such that the final objective function is
\begin{equation}
J =\kern-5mm
\sum_{(x,y) \in \Csource}\kern1mm \sum_{(s,t) \in L}\kern0mm
\alpha \cdot H(x, y) + (1 - \alpha) \cdot || M \cdot s - M' \cdot t ||
\end{equation}
thereby simplifying the model and removing the $P$ parameter. Table \ref{acl:noproj} shows that \blse outperforms this simplified model on all tasks. 

\begin{table}[]
\centering
\renewcommand*{\arraystretch}{0.8}
\setlength\tabcolsep{2.0mm}
\begin{tabular}{llrrr}
\toprule
&& \blse & no proj. \\ 
\cmidrule(l){3-3}\cmidrule(l){4-4}
\multirow{3}{*}{\rt{binary}}  & ES & 74.6 & 52.0 \\
&CA &  72.9 & 48.3\\
&EU &  69.3 & 49.1\\
\cmidrule(l){3-3}\cmidrule(l){4-4}

\multirow{3}{*}{\rt{4-class}}  &ES & 41.2 & 21.3 \\
&CA & 35.9  & 18.3 \\
&EU & 30.0  & 17.0\\
\bottomrule
\end{tabular}
\caption[No projection]{An empirical comparison of \blse and a simplified model which directly projects the embeddings to the sentiment classes. \blse outperforms the simplified model on all tasks.}
\label{acl:noproj}
\end{table}

\subsubsection{Qualitative Analyses of Joint Bilingual Sentiment Space}
\label{qualitativeanalysis}

In order to understand how well our model transfers sentiment
information to the target language, we perform two qualitative
analyses. First, we collect two sets of 100
positive sentiment words and one set of 100 negative sentiment
words. An effective cross-lingual sentiment classifier using embeddings should learn 
that two positive words should be closer in the shared bilingual 
space than a positive word
and a negative word. We test if \blse is able to do this by training our model and
after every epoch observing the mean cosine similarity between the
sentiment synonyms and sentiment antonyms after projecting to the
joint space.

\begin{figure*}[]
\begin{center}
\includegraphics[width = 0.9\linewidth]{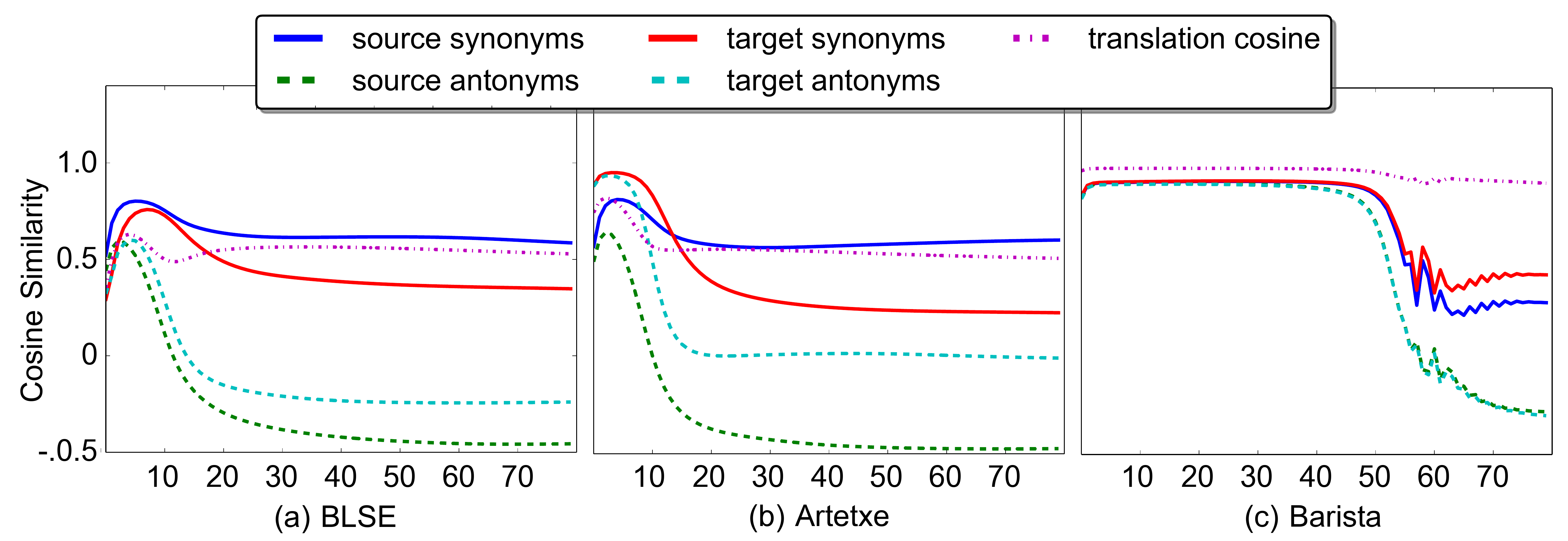}
\end{center}
\caption[Analysis of separation of classes with \blse]{Average cosine similarity between a subsample of translation
  pairs of same polarity (``sentiment synonyms'') and of opposing
  polarity (``sentiment antonyms'') in both target and source
  languages in each model. The x-axis shows
  training epochs. We see that \blse is able to learn that sentiment
  synonyms should be close to one another in vector space and
  sentiment antonyms should not.
}
\label{acl:fig:synant}
\end{figure*}

We compare \blse with \vecmap and \barista by replacing the Linear
SVM classifiers with the same multi-layer classifier used in \blse and
observing the distances in the hidden layer. Figure~\ref{acl:fig:synant}
shows this similarity in both source and target language, along with
the mean cosine similarity between a held-out set of translation pairs
and the macro \F scores on the development set for both source and
target languages for \blse, \barista, and \vecmap. From this plot, it
is clear that \blse is able to learn that sentiment synonyms should be
close to one another in vector space and antonyms should have a
negative cosine similarity. While the other models also learn this to
some degree, jointly optimizing both sentiment and projection gives
better results.

Secondly, we would like to know how well the projected vectors compare
to the original space. Our hypothesis is that some relatedness and
similarity information is lost during projection. Therefore, we
visualize six categories of words in t-SNE, which projects high dimensional representations to lower dimensional spaces while preserving the relationships as best as possible \shortcite{Vandermaaten2008}:
positive sentiment words, negative sentiment words, functional words,
verbs, animals, and
transport. 

\begin{figure}[]
\centering
\includegraphics[width = 0.4\linewidth]{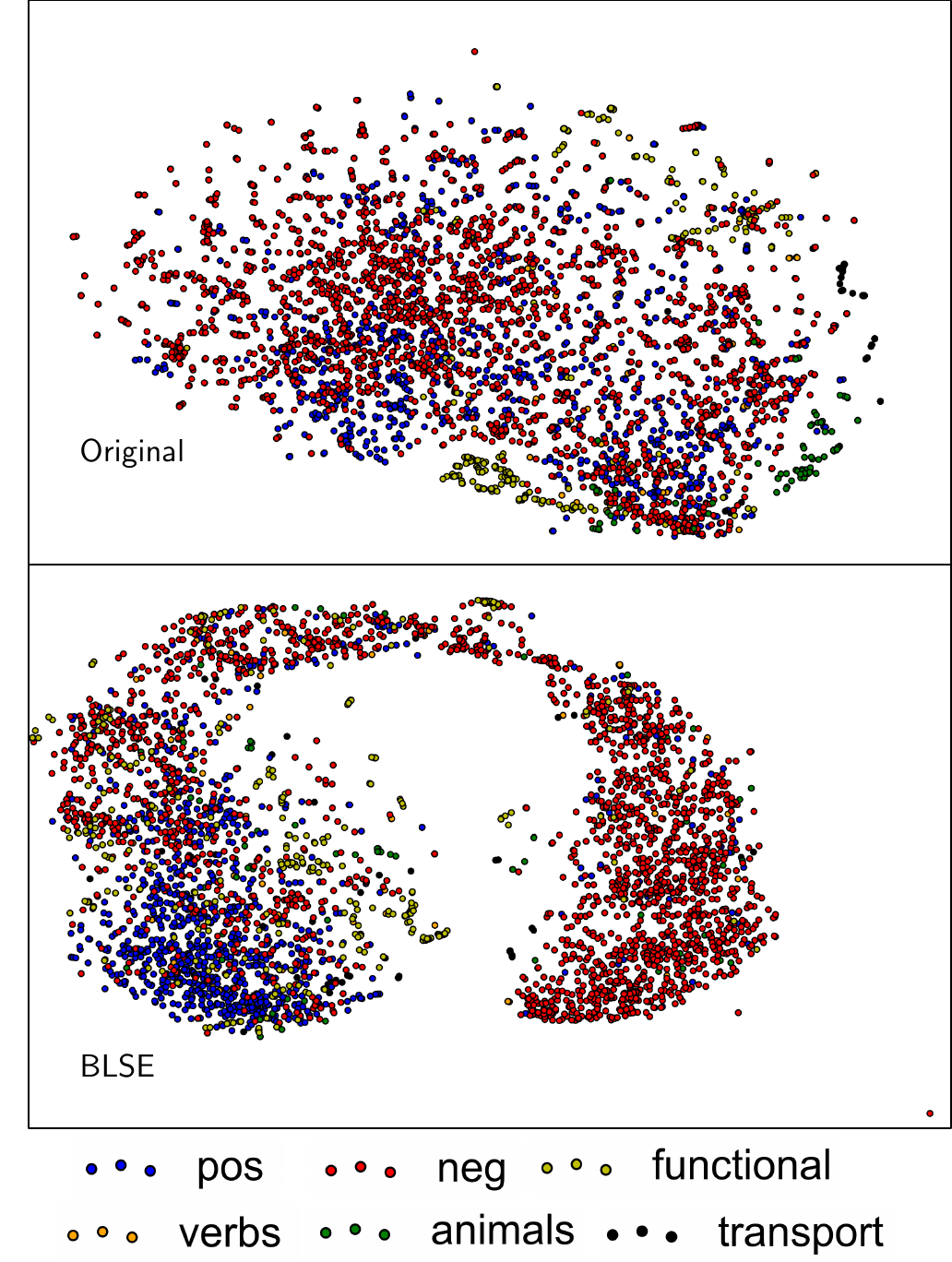}
\caption[t-SNE visualization of \blse]{t-SNE-based visualization of the Spanish vector space before
  and after projection with \blse. There is a clear separation of
  positive and negative words after projection, despite the fact that
  we have used no labeled data in Spanish.}
\label{acl:fig:tsne}
\end{figure}

The t-SNE plots in Figure \ref{acl:fig:tsne} show that the positive and
negative sentiment words are rather clearly separated after projection
in \blse. This indicates that we are able to incorporate sentiment
information into our target language without any labeled data in the
target language. However, the downside of this is that functional
words and transportation words are highly correlated with positive
sentiment.

\subsubsection{Analysis of $\alpha$ parameter}

\begin{figure}[]
\centering
\includegraphics[width = 0.6\linewidth]{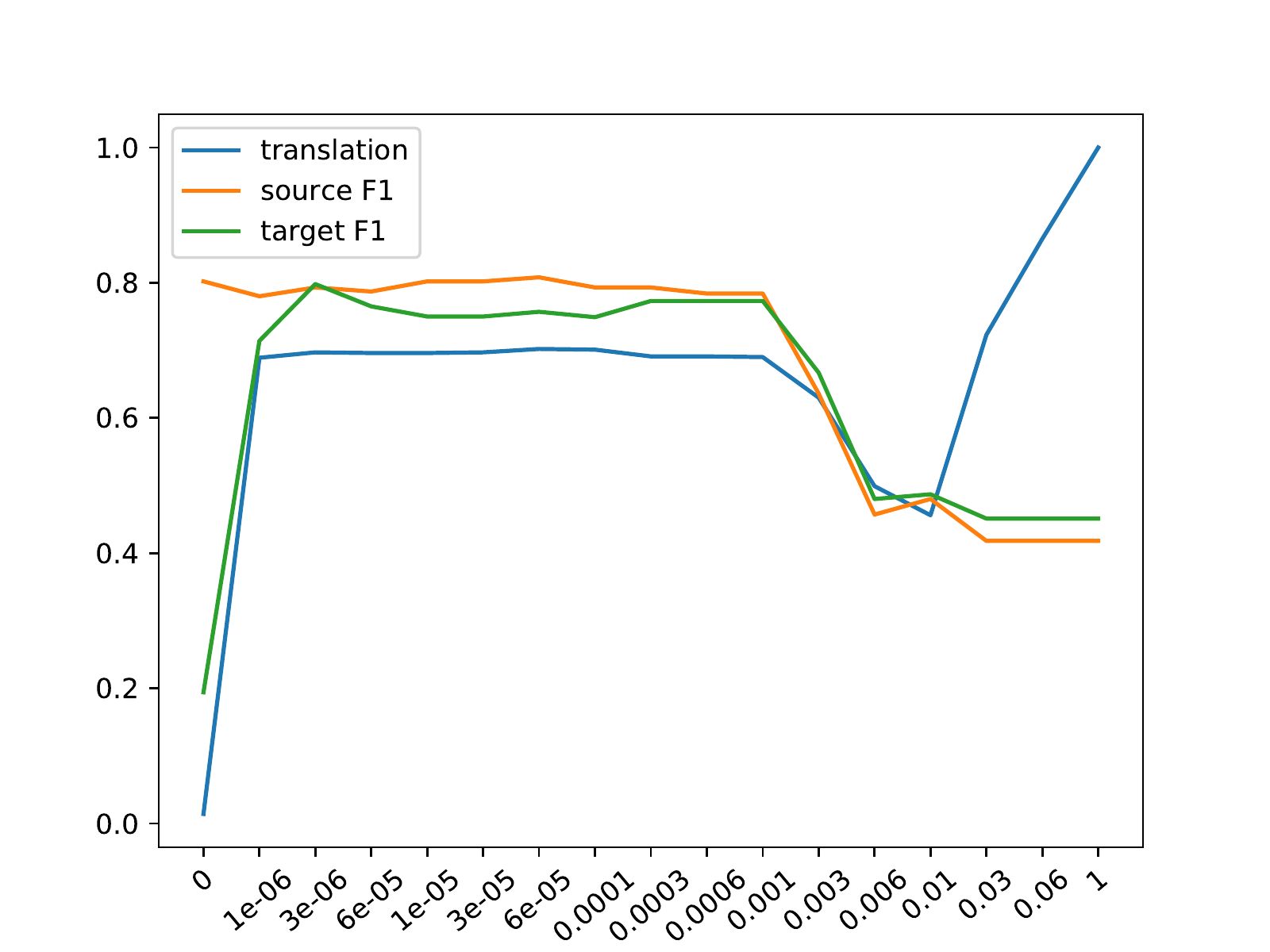}
\caption[Analysis of $\alpha$]{An analysis of the $\alpha$ parameter of \blse showing cosine similarity of translation pairs and macro \F for source and target development data. The optimal values range from $1\times10^{-6}$ to $1\times10^{-3}$.}
\label{acl:fig:alpha}
\end{figure}

Finally, in order to analyze the sensitivity of the alpha parameter, we train \blse models for 30 epochs each with $\alpha$ between 0 and 1. Figure \ref{acl:fig:alpha} shows the average cosine similarity for the translation pairs, as well as macro \F for both source and target language development data.

Values near $0$ lead to poor translation and consecuently poor target language transfer. There is a rather large ``sweet spot'' where all measures perform best and finally, the translation is optimized to the detriment of sentiment prediction in both source and target languages with values near $1$.

\subsubsection{Discussion}

The experiments in this section have proven that it is possible to
perform cross-lingual sentiment analysis without machine translation,
and that jointly learning to project and predict sentiment is
advantageous. This supports the growing trend of jointly training for
multiple objectives \shortcite{Tang2014,Klinger2015,Ferreira2016}.

This approach has also been exploited within the framework of
multi-task learning, where a model learns to perform multiple similar
tasks in order to improve on a final task
\shortcite{Collobert2011a}. The main difference between the joint
method proposed here and multi-task learning is that vector space
projection and sentiment classification are not similar enough tasks
to help each other. In fact, these two objectives compete against one
another, as a perfect projection would not contain enough information
for sentiment classification, and vice versa.

\subsection{Experiment 2: Targeted Classification}
\begin{table*}
\newcommand{\sepp}{\cmidrule(r){3-3}\cmidrule(r){4-4}\cmidrule(r){5-6}\cmidrule(r){7-7}\cmidrule(r){8-8}\cmidrule(r){9-9}}
\newcommand{\seppp}{\cmidrule(r){1-2}\cmidrule(r){3-3}\cmidrule(r){4-4}\cmidrule(r){5-6}\cmidrule(r){7-7}\cmidrule(r){8-8}\cmidrule(r){9-9}}
\newcommand{\sep}{\cmidrule(r){2-2}\cmidrule(r){3-3}\cmidrule(r){4-4}\cmidrule(r){5-6}\cmidrule(r){7-7}\cmidrule(r){8-8}\cmidrule(r){9-9}}
\definecolor{green}{RGB}{150,255,150}
\definecolor{blue}{RGB}{150,150,255}
\newcommand{\bestproj}[1]{{\setlength{\fboxsep}{0pt}\colorbox{lightblue}{\textit{#1}}}}
\newcommand{\bestoverall}[1]{{\setlength{\fboxsep}{0pt}\colorbox{lightgreen}{\textbf{#1}}}}
\setlength\tabcolsep{10pt}
\renewcommand*{\arraystretch}{0.8}
\centering\small
\begin{tabular}{lll|ccccc|c}
\toprule
& & & EN-ES & EN-CA & EN-EU & EN-ES & EN-DE \\
& && \multicolumn{1}{c}{OpeNER}&\multicolumn{2}{c}{MultiBooked}&\multicolumn{1}{c}{SemEval}&\multicolumn{1}{c}{USAGE}&\multicolumn{1}{c}{Average}\\
\sepp
\multirow{24}{*}{\rt{Binary}} & \multirow{6}{*}{\rt{\sent}} 
& \blse & 64.4 & 47.3 & 45.5 & 61.1 & 63.8 & 56.4 \\ 
&& \vecmap & 52.2 & 41.8 & 39.1 & 42.3 & 31.2 & 51.3\\
&& \muse & 47.6 & 40.1 & 45.8 & 45.3 & 47.5 & 45.3\\
&& \barista & 47.3 & 39.1 & 45.8 & 42.3 & 33.4 & 41.6\\
&& \mt & 70.8 & \bestoverall{81.5} &
\bestoverall{76.2} & 70.9 & 58.8 & 71.6 \\
&& \unsup & \bestoverall{73.7} & 69.8 & 70.1 & 66.1 & - & - \\
\sep
& \multirow{6}{*}{\rt{\splitsent}}
& \blse & \bestproj{66.8} & \bestproj{69.8} & \bestproj{66.3} & \bestproj{62.2} & 50.0 & \bestproj{63.0}\\ 
&& \vecmap & 65.8 & 64.4 & 65.1 & 60.0 & 39.9 & 59.0\\
&& \muse & 58.3 & 64.3 & 50.2 & 59.8 & \bestproj{57.0} & 57.9 \\
&& \barista & 61.9 & 59.0 & 56.1 & 44.5 & 35.3 & 51.4 \\
&& \mt & 67.3 & 77.8 & 74.8 & 73.2 & 69.4 & 72.5\\
&& \unsup & 71.6 & 73.5 & 64.0 & \bestoverall{77.1} & - & - \\
\sep
&\multirow{5}{*}{\rt{\cononlytable}}
& \blse & 47.3 & 39.1 & 45.8 & 42.3 & 55.9 & 46.1\\ 
&& \vecmap & 47.3 & 39.1 & 45.8 & 42.3 & 45.8 & 44.1 \\
&& \muse & 55.5 & 67.5 & 52.1 & 61.6 & 45.4 & 56.4 \\
&& \barista & 47.3 & 60.2 & 51.9 & 42.3 & 45.5 & 49.4 \\
&& \mt & 66.5 & 78.1 & 72.4 & 74.2 & \bestoverall{73.1} & \bestoverall{72.9}\\
&& \unsup & 69.9 & 72.3 & 63.6 & 75.5 & - &-  \\
\sep
&\multirow{6}{*}{\rt{\asponlytable}}
& \blse &  53.1 & 43.7 & 42.7 & 42.3 & 41.5 & 44.7\\ 
&& \vecmap & 54.4 & 51.1 & 35.4 & 45.5 & 45.2 & 46.3\\
&& \muse &   56.2 & 55.4 &  52.3 & 46.0 & 47.5  &  51.5  \\
&& \barista & 48.9 & 53.0 & 48.5 & 42.3 & 44.8 & 47.5\\
&& \mt & 46.7 & 40.1 & 45.8 & 47.5 & 56.0 & 47.2\\
&& \unsup & 52.4 & 51.0 & 52.1 & 43.7 & - &- \\
\sep
&\multirow{1}{*}{Maj.} & &  47.3 & 39.1 & 45.8 & 42.3 & 43.0 & 43.5\\
\seppp
\multirow{24}{*}{\rt{Multiclass}} & \multirow{6}{*}{\rt{\sent}} 
& \blse & 25.2 & 23.3 & 16.6 & 36.0 & \bestoverall{40.5} & 28.3\\ 
&& \vecmap & 28.1 & 19.9 & 26.3 & 28.2 & 28.3 & 26.2\\
&& \muse & 22.4 & 23.2 & 23.5 & 27.4 & 24.1 & 24.1\\
&& \barista & 29.3 & 35.8 & 27.0 & 27.4 & 29.9 & 29.9\\
&& \mt & 41.4 & \bestoverall{46.5} &
\bestoverall{44.3} & 33.1 & 28.9 & \bestoverall{38.8}\\
&& \unsup & \bestoverall{42.7} & 37.7 & 37.6 & 32.5 & - & -\\
\sep
& \multirow{6}{*}{\rt{\splitsent}}
& \blse & 18.5 & 14.3 & 15.7 & \bestoverall{40.6} & 29.5 & 23.7\\ 
&& \vecmap & 29.2 & 30.9 & 28.0 & 38.9 & 27.9 & 31.0\\
&& \muse & \bestproj{32.9} & \bestproj{33.5} & \bestproj{27.3} & 27.4 & 39.7 & \bestproj{32.2}\\
&& \barista & 27.9 & 35.1 & 27.3 & 27.4 & 33.4 & 28.1\\
&& \mt & 24.7 & 29.2 & 27.0 & 33.8 & 33.2 & 29.6\\
&& \unsup & 28.9 & 26.9 & 23.9 & 31.7 & - & -  \\
\sep
&\multirow{6}{*}{\rt{\cononlytable}}
& \blse & 18.5 & 12.6 & 15.7 & 27.4 & 38.4 & 22.5\\ 
&& \vecmap & 18.5 & 12.6 & 15.7 & 27.4 & 28.3 & 20.5\\
&& \muse & 22.7 & 39.0 & 27.4 & 27.4 & 30.0 & 29.3\\
&& \barista & 32.9 & 31.6 & 27.2 & 27.4 & 32.1 & 30.2\\
&& \mt & 27.5 & 31.4 & 27.2 & 30.6 & 34.4 & 30.2\\
&& \unsup &  29.4 & 27.7 & 23.9 & 32.8 & - & -\\
\sep
&\multirow{6}{*}{\rt{\asponlytable}}
& \blse &  19.1 & 17.3 & 16.7 & 27.4 & 25.3 & 21.2\\ 
&& \vecmap &  25.8 & 23.1 & 19.0 & 32.1 & 25.3 & 25.1\\
&& \muse &   23.2 &  21.6 & 17.1 & 29.5 & 31.1 & 24.5\\
&& \barista & 21.8 & 21.5 & 16.8 & 27.4 & 33.9 & 24.3 \\
&& \mt & 26.9 & 23.3 & 23.9 & 30.5 & 33.6 & 27.6 \\
&& \unsup & 22.9 & 18.7 & 21.2 & 19.7 & - &-  \\
\sep
&\multirow{1}{*}{Maj.} & &  18.5 & 12.6 & 15.7 & 27.4 & 28.3 & 20.5\\
\bottomrule
\end{tabular}
\caption{Macro \F results for all corpora and techniques. We denote
  the best performing projection-based
  method per column with a \bestproj{blue box} and the best overall method
  per column with a
  \bestoverall{green box}.}
\label{results:all}
\end{table*}

Table \ref{results:all} shows the macro \F scores for all
cross-lingual approaches (\blse, \vecmap, \muse, \barista, \mt, \unsup) and all
targeted approaches (\sent, \splitsent, \cononly, and \asponly). The final column is the average over all
corpora. The final row in each setup shows the macro \F for a
classifier that always chooses the majority class.

\blse outperforms other projection methods on the binary setup, 63.0
macro averaged \F across corpora versus 59.0, 57.9, and 51.4 for
\vecmap, \muse, and \barista, respectively. On the multiclass setup,
however, \muse (32.2 \F) is the best, followed by \vecmap (31.0),
\barista (28.1) and \blse (23.7). 
\unsup performs well across all experiments,
achieving the best results on OpeNER ES (73.2 on binary and 42.7 on multiclass) and SemEval binary (77.1). \vecmap is never the best nor the
worst approach. In general, \barista performs poorly on the binary
setup, but slightly better on the multiclass, although the overall
performance is still weak. These results are similar to those observed
in Experiment 1 for sentence classification.

The \splitsent approach to ABSA improves over the \sent baseline
on 33 of 50 experiments, especially on binary (21/25), while on
multiclass it is less helpful (13/25). Both \sent and \splitsent
 normally outperform \cononly
or \asponly approaches. This confirms the intuition that it is
important to take both context and target information for
classification. Additionally, the \cononly approach always 
performs better than \asponly,
which indicates that context is more important than the prior probability
of an target being positive or negative.

Unlike the projection methods, \mt using only the \sent representation 
performs well on the OpeNER and MultiBooked datasets,
while suffering more on the SemEval~and USAGE datasets. This is explained by
the percentage of sentences that contain contrasting polarities in
each dataset: between 8 and 12\% for the OpeNER and Multibooked datasets,
compared to 29\% for SemEval or 50\% for USAGE. In sentences with
multiple contrasting polarities, the \sent baseline performs poorly.

Finally, the general level of performance of projection-based targeted
cross-lingual sentiment classification systems shows that they still
lag 10+ percentage points behind \mt on binary (compare \mt (72.9 \F)
with \blse (63.0)), and 6+ percentage points on multiclass (\mt (38.8)
versus \muse (32.2)). The gap between \mt and projection-based
approaches is therefore larger on targeted sentiment analysis than at
sentence-level.

\subsubsection{Error Analysis}
\label{erroranalysis}

\begin{table}
\setlength\tabcolsep{10pt}
\renewcommand*{\arraystretch}{0.8}
\centering\small
\begin{tabular}{lcc}
\toprule
& correct & incorrect \\
\cmidrule(lr){2-2}\cmidrule(lr){3-3}
\blse & 2.1 & 2.5 \\
\vecmap & 2.5 & 2.1 \\
\muse & 2.1 & 2.2 \\
\barista & 1.7 & 2.2 \\
\mt & 2.1 & 2.2 \\
\unsup & 2.1 & 2.2 \\
\bottomrule
\end{tabular}
\caption{Average length of tokens of correctly and incorrectly classified
targets on the OpeNER Spanish binary corpus.}
\label{lengths}
\end{table}

We perform a manual analysis of the targets misclassified by all systems
on the OpeNER Spanish binary corpus (see Table \ref{lengths}),
and found that the average length of misclassified targets
is slightly higher than that of correctly classified targets, except
for with \vecmap. This indicates that averaging may have a detrimental effect as
the size of the targets increases.

With the \mt upperbounds, there is a non-negligible amount of noise
introduced by targets which have been incorrectly translated
(0.05\% OpeNER ES, 6\% MultiBooked EU, 2\% CA, 2.5\% SemEval, 1\% USAGE). We hypothesize that
this is why \mt with \cononly performs better than \mt with \splitsent. This
motivates further research with projection-based methods, as they do
not suffer from translation errors.

\begin{figure*}
  \centering
  \includegraphics[width=1\textwidth]{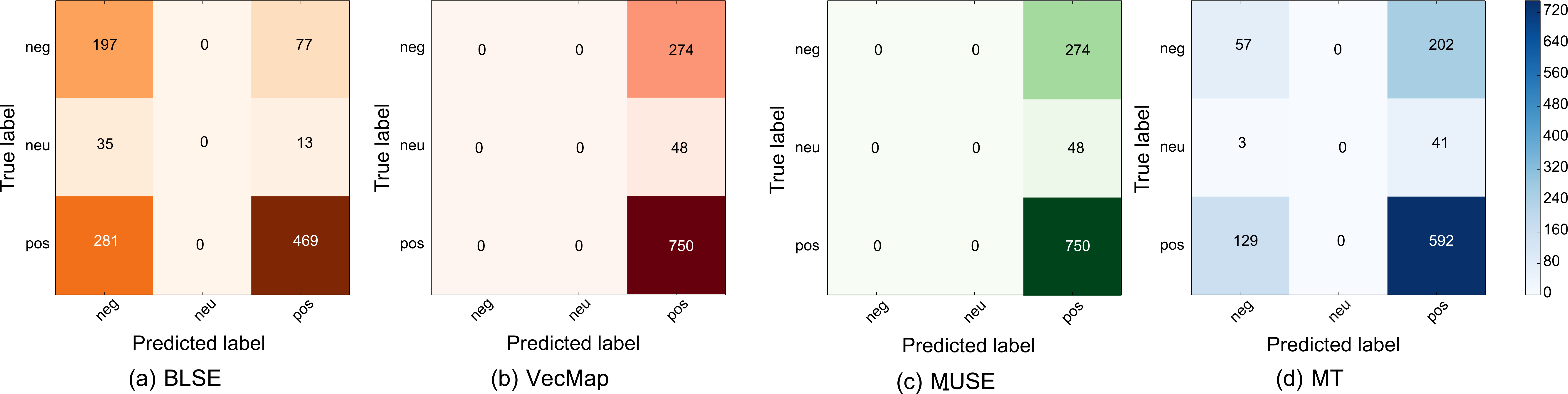}
  \caption{Confusion matrices for all \splitsent models on the SemEval task.}
  \label{fig:confusionmatrix}
\end{figure*}

The confusion matrices of the models on the SemEval task, shown in Figure \ref{fig:confusionmatrix}, show that on the multilabel task, models are not able
to learn the neutral class. This derives from
the large class imbalance found in the data (see Table
\ref{stats:datasets}). Similarly, models do not learn the 
Strong Negative class on the OpeNER and MultiBooked datasets.

\section{Case Study: Real World Deployment}
\label{sec:deployment}

\subsection{Motivation}
The performance of machine learning models on different target
languages depends on the amount of data available, the quality of the
data, and characteristics of the target language, \eg, morphological
complexity. In the following, we analyze these aspects. There has been
previous work that has observed target-language specific differences
in multilingual dependency parsing \shortcite{Zeljko2016}, machine
translation \shortcite{Johnson2017}, and language modeling
\shortcite{Cotterell2018,Gerz2018}. We are not aware of any work in
cross-lingual sentiment analysis that explores the relationship
between target language and performance in such depth and aim at
improving this situation in the following.

Additionally, the effect of domain differences when performing
cross-lingual tasks has not been studied in
depth. \shortciteA{Hangya2018} propose domain adaptation methods for
cross-lingual sentiment classification and bilingual dictionary
induction. They show that creating domain-specific cross-lingual
embeddings improves the classification for English-Spanish. However,
the source-language training data used to train the sentiment
classifier is taken from the same domain as the target-language test
data. Therefore, it is not clear what the effect of using
source-language training data from different domains would be. We
analyzed the model presented in Section~\ref{section:blse} in a domain
adaptation setup, including the impact of domain differences
\shortcite{Barnes2018c}. The main result was that our model performs
particularly well on more distant domains, while other approaches
\shortcite{Chen2012,Ziser2017} performed better when the source and
target domains were not too dissimilar.

In the following, we transfer this analysis to the target-based
projection model in a real-world case study which mimics a user
searching for the sentiment on touristic attractions. In
order to analyze how well these methods generalize to new languages
and domains, we deploy the
targeted \blse, \muse, \vecmap and \mt models on tweets in ten Western
European languages with training data from three different domains. Additionally, we include experiments with the \unsup models for a subset of the languages.
English is the source language in all experiments,
and we test on each of the ten target languages and attempt to answer the
following research questions:

\begin{itemize}
\item How much does the amount of monolingual data available to create
  the original embeddings effect the final results?
\item How do features of the target language, \ie similarity to source
  language or morphological complexity, affect the performance?
\item How do domain mismatches between source-language training and
  target-language test data affect the performance?
\end{itemize}

Section \ref{deployment:erroranlysis} addresses our findings regarding 
these questions  and demonstrates that 1) the amount of
monolingual data does not correlate with classification results, 2)
language similarity between the source and target languages based on
word and character n-gram distributions predicts the performance of
\blse on new datasets, and 3) domain mismatch has more of an effect on
the multiclass setup than binary.

\subsection{Experimental Setup}
%
\subsubsection{Setup}
\label{casestudysetup}
%
%

We collect tweets directed at a number of
tourist attractions in European cities using the Twitter API in 10
European languages, including several under-resourced languages
(English, Basque, Catalan, Galician, French, Italian, Dutch, German,
Danish, Swedish, and Norwegian). We detail the data collection and
annotation procedures in Section \ref{deployment:datacollection}. For
classification, we compare \mt the best performing projection-based
methods (\blse, \muse, \vecmap) using the \splitsent method, detailed
further in Section \ref{deployment:experiments}. As we need
monolingual embeddings for all projection-based approaches, we create
skipgram embeddings from Wikipedia dumps, detailed in Section
\ref{deployment:embeddings}.

\subsubsection{Data Collection}
\label{deployment:datacollection}

As an experimental setting to measure the effectiveness of targeted
cross-lingual sentiment models on a large number of languages, we
collect and annotate small datasets from Twitter for each of the
target languages, as well as a larger dataset to train the models in
English. While it would be possible to only concentrate our efforts on
languages with existing datasets in order to enable evaluation, this
could give a distorted view of how well these models generalize. In
order to reduce the possible ambiguity of the tourist attractions, we
do not include those that have two or more obvious senses, \eg,
Barcelona could refer either to the city or the football team.

\begin{table}[t]
\begin{center}
\centering\small
\renewcommand*{\arraystretch}{1.0}
\setlength\tabcolsep{1.1mm}
\begin{tabularx}{\linewidth}{lXl}
\toprule
Canonical name & Search terms & City \\
\cmidrule(r){1-1}\cmidrule(rl){2-2}\cmidrule(l){3-3}
The Sagrada Familia Church & sagrada familia & Barcelona \\
Güell Park & parc güell & Barcelona \\
La Boqueria Market & la boqueria & Barcelona \\
Tibidabo Theme Park & tibidabo & Barcelona \\
Santiago de Compostela & santiago de compostela & Santiago de Compostela \\
The Guggenheim Museum Bilbao & guggenheim bilbao & Bilbao \\
Txindoki Mountain & txindoki & Basque Country \\
Anboto Mountain & anboto & Basque Country \\
The Eiffel Tower & tour eiffel; eiffel tower; torre eiffel; eiffel
dorrea; eiffelturm; eiffeltårnet; eiffeltornet; eiffeltoren & Paris \\
The Louvre museum & louvre & Paris \\
The Champs-Élysées & champs-élysées & Paris \\
Big Ben tower & big ben & London \\
The London Eye & london eye & London \\
Buckingham Palace & buckingham palace; palacio buckingham; palau
buckingham; jauregia buckingham & London \\
Akershus Castle Oslo & akerhus slot; akerhus fortress; fortaleza
akershus; fortezza akershus; fortalesa akershus; akershus gotorlekua;
festung akershus; akershus fästning  & Oslo \\
The Oslo Viking Ship Museum & vikingskipshuset oslo; oslo viking ship
museum; museo de barcos vikingos oslo; museu de vaixells víkings Oslo;
oslo itsasontzi bikingoen museoa; Musée des navires vikings Oslo& Oslo \\
The Gamla Stan Stockholm    & gamla stan stockholm; gamla stan
estocolmo; gamla stan estocolm & Stockholm \\
\bottomrule
\end{tabularx}
\end{center}
\caption[Tourist targets]{Touristic targets used as tweet search
  criteria.}
\label{deployment:table:targets}
\end{table}

In order to obtain a varied sample of tweets with subjective opinions,
we download tweets that contain mentions of these tourist attractions
as well as one of several emoticons or keywords\footnote{The emoticons
  and keywords we used were ``:)'', ``:('', ``good'', ``bad'', and the
  translations of these last two words into each target
  language.}. This distant supervision technique has been used to
create sentiment lexicons \shortcite{Mohammad2016}, semi-supervised
training data \shortcite{Felbo2017}, and features for a classifier
\shortcite{Turney2003}. We then remove any tweets that are less than 7
words long or which contain more than 3 hashtags or mentions. This
increases the probability that a tweet text contains sufficient
information for our use case setting.

We manually annotate all tweets for its polarity toward the target to
insure the quality of the data\footnote{Data is available at\\ \url{https://github.com/jbarnesspain/targeted_blse/tree/master/case_study/datasets}}. Note that we only annotate the sentiment towards the
predefined list of targets, which leads to a single annotated target
per tweet. Any tweets that have unclear polarity towards the target
are assigned a neutral label. This produces the three class setup that
is commonly used in the SemEval tasks
\shortcite{Nakov2013,Nakov2016}. Annotators were master's and doctoral
students between 27 and 35 years old. All had either native or C1
level fluency in the languages of interest. Finally, for a subset of
tweets in English, Catalan, and Basque two annotators classify each
tweet. Table~\ref{deployment:table:examples} shows three example
tweets from English.

\begin{table*}[]
\newcommand{\sepp}{\cmidrule(lr){2-2}\cmidrule(lr){3-3}\cmidrule(lr){4-4}\cmidrule(lr){5-5}\cmidrule(lr){6-6}\cmidrule(lr){7-7}\cmidrule(lr){8-8}\cmidrule(lr){9-9}\cmidrule(lr){10-10}\cmidrule(lr){11-11}\cmidrule(lr){12-12}}
\centering\small
\renewcommand*{\arraystretch}{1.0}
\setlength\tabcolsep{1.8mm}
\begin{tabular}{cccccccccccc}
\toprule
    & EN  & EU & CA  & GL & IT  & FR  & NL & DE  & NO & SV & DA \\
\sepp
$+$ & 388 & 40 & 88  & 27 & 63  & 51  & 30 & 72  & 47 & 40 & 34 \\
0   & 645 & 93 & 165 & 57 & 103 & 140 & 48 & 125 & 80 & 93 & 77 \\
$-$ & 251 & 9  & 47  & 15 & 56  & 66  & 8  & 48  & 10 & 20 & 11 \\
\sepp
IAA & 0.62 & 0.60 & 0.61 & --- & --- & --- & --- & --- & --- & --- & --- \\ 
\bottomrule
\end{tabular}
\caption[Deployment dataset statistics]{Statistics of Tweet corpora collected for the deployment study, as well as inter-annotator agreement for English, Basque, and Catalan calculated with Cohen's $\kappa$.}
\label{deployment:table:tweets}
\end{table*}

Table \ref{deployment:table:tweets} depicts the number of annotated
targets for all languages, as well as inter-annotator agreement using
Cohen's $\kappa$.  The neutral class is the largest in all languages,
followed by positive, and negative. These distributions are similar to
those found in other Twitter crawled datasets
\shortcite{Nakov2013,Nakov2016}. We calculate pairwise agreement on a
subset of languages using Cohen's $\kappa$. The scores reflect a good
level of agreement (0.62, 0.60, and 0.61 for English, Basque, and
Catalan, respectively).

\begin{table}
\centering
\begin{tabular}{ll}
\toprule
Text & Label \\
\cmidrule(r){1-1}\cmidrule(l){2-2}
I'm so jealous. I want to visit the \underline{Sagrada Familia}! & Positive \\

just visited yoko ono's works at the bilbao \underline{guggenheim museum} & Neutral \\

low points: I visited the \underline{Buckingham Palace} & Negative \\
\bottomrule
\end{tabular}
\caption{Three example tweets in English. The underlined phrases are the targets.}
\label{deployment:table:examples}
\end{table}

\subsubsection{Embeddings}
\label{deployment:embeddings}

We collect Wikipedia dumps for ten languages; namely, Basque, Catalan, Galician, French, Italian, Dutch, German, Danish, Swedish, and Norwegian. We then preprocess them using the Wikiextractor script\footnote{\url{http://attardi.github.io/wikiextractor/}}, and sentence and word tokenize them with either IXA pipes \shortcite{Agerri2014} (Basque, Galician, Italian, Dutch, and French), Freeling \shortcite{Padro2010}
 (Catalan), or NLTK \shortcite{Loper2002}
 (Norwegian, Swedish, Danish).

For each language we create Skip-gram embeddings with the word2vec toolkit following the pipeline and parameters described in Section \ref{embeddings}. This process gives us 300 dimensional vectors trained on similar data for all languages. We assume that any large differences in the embedding spaces derive from the size of the data and the characteristics of the language itself. Following the same criteria laid out in Section \ref{transdict}, we create projection dictionaries by translating the Hu and Liu dictionary \shortcite{HuandLiu2004} to each of the target languages and keeping only translations that are single word to single word. The statistics of all Wikipedia corpora, embeddings, and projection dictionaries are shown in Table \ref{deployment:table:stats}.

\begin{table*}[]
\newcommand{\sepp}{\cmidrule(r){1-1}\cmidrule(lr){2-2}\cmidrule(lr){3-3}\cmidrule(lr){4-4}\cmidrule(lr){5-5}\cmidrule(lr){6-6}\cmidrule(lr){7-7}\cmidrule(lr){8-8}\cmidrule(lr){9-9}\cmidrule(lr){10-10}\cmidrule(lr){11-11}\cmidrule(lr){12-12}}
\centering\small
\renewcommand*{\arraystretch}{1.0}
\setlength\tabcolsep{1.1mm}
\begin{tabular}{llrrrrrrrrrr}
\toprule
Type & Measurement  & EU & CA & GL & IT & FR & NL & DE & NO & SV & DA \\
\sepp
\multirow{2}{*}{Wiki}
&  sents. (M) & 3.1  & 9.6 & 2.5  & 23.7  & 39.1  & 19.4  & 53.7  & 6.8  & 35.9   & 3.6\\
&  tokens (M) & 47.9 & 143.7& 51.0 & 519.6 & 771.8 & 327.3 & 902,1 & 110.5 & 457.3 & 64.4\\
\sepp
\multirow{2}{*}{Emb}
&  vocab. (k) & 246.0 &400.9& 178.6 & 729.4 & 967.7 & 877.9 & 2,102.7 & 443.3 & 1,346.7 & 294.6\\
&  dimension & 300 & 300 & 300 & 300 & 300 & 300 & 300 & 300 & 300 & 300\\
\sepp
\multirow{1}{*}{Dict}
&  pairs & 4,616 & 5,271 & 6,297 & 5,683 & 5,383 & 5,700 & 6,391 & 5,177 & 5,344 & 5,007\\    
\bottomrule
\end{tabular}
\caption{Statistics of Wikipedia corpora, embeddings, and projection
  dictionaries (M denotes million, k denotes thousand).}
\label{deployment:table:stats}
\end{table*}

\subsubsection{Experiments}
\label{deployment:experiments}

Since we predetermine the sentiment target for each tweet, we can perform targeted experiments without further annotation. We use the \splitsent models described in Section \ref{targetedprojection}. Our model is the  targeted \blse models described in Section \ref{targetedprojection}. Additionally, we compare to the targeted \muse, \vecmap, and \mt models, as well as an Ensemble classifier that uses the predictions from \blse and \mt before taking the largest predicted class for classification (see Section \ref{experiment1} for details). Finally, we set a majority baseline by assigning the most common label (neutral) to all predictions. All models are trained for 300 epochs with a learning rate of 0.001 and $\alpha$ of 0.3. 

We train the five models on the English data compiled during this study, as well as on the USAGE, and SemEval English data (the details can be found in Table \ref{stats:datasets}) and test the models on the target-language test set.

\subsection{Results}
\label{deployment:results}

\begin{small}
\begin{table}

\newcommand{\sep}{\cmidrule(lr){4-4}\cmidrule(lr){5-5}\cmidrule(lr){6-6}\cmidrule(lr){7-7}\cmidrule(lr){8-8}\cmidrule(lr){9-9}\cmidrule(lr){10-10}\cmidrule(lr){11-11}\cmidrule(lr){12-12}\cmidrule(lr){13-13}}

\centering\small
\renewcommand*{\arraystretch}{0.8}
\setlength\tabcolsep{1.1mm}
\begin{tabular}{lllcccccccccc|c}
\toprule
\multirow{19}{*}{\rt{Binary}} &
Training Data & Model & EU & CA & GL & IT & FR & NL & DE & NO & SV & DA & avg. \\
\cmidrule(r){2-2}\cmidrule(r){3-3}\cmidrule(lr){4-4}\cmidrule(lr){5-5}\cmidrule(lr){6-6}\cmidrule(lr){7-7}\cmidrule(lr){8-8}\cmidrule(lr){9-9}\cmidrule(lr){10-10}\cmidrule(lr){11-11}\cmidrule(lr){12-12}\cmidrule(lr){13-13}
& & maj. class &  46.0 & 39.5 & 38.8 & 34.6 & 36.1 & 44.1 & 37.5 & 43.4 & 45.2 & 40.0 & 40.5\\
\cmidrule(r){3-3}\cmidrule(lr){4-4}\cmidrule(lr){5-5}\cmidrule(lr){6-6}\cmidrule(lr){7-7}\cmidrule(lr){8-8}\cmidrule(lr){9-9}\cmidrule(lr){10-10}\cmidrule(lr){11-11}\cmidrule(lr){12-12}\cmidrule(lr){13-13}
&\multirow{6}{*}{Twitter}
& \blse &  53.7 & 54.5 & 52.0 & \textbf{63.4} & 49.2 & 44.1 & 53.4 & 56.4 & \textbf{65.3} & \textbf{68.3} & \textbf{56.0}\\
&&\muse & 53.7 & 50.0 & 55.9 & 49.5 & 40.6 & 18.3 & 51.5 & 47.9 & 52.0 & 67.4 & 48.7\\
&&\vecmap &  56.4 & 48.1 & 33.3 & 38.2 & 48.6 & 55.2 & 51.0 & 59.0 & 60.5 & 43.4 & 49.4\\ 
&&\mt &  41.3 & 41.4 & \textbf{56.5} & 39.7 & 54.5 & 43.3 & 55.1 & 52.2 & 49.8 & 55.6 & 48.9 \\
&& \unsup & 44.7 & 47.3 & - & - & - & - & - & - & - & - & -\\
&& Ensemble & 40.5 & 42.5 & 41.8 & 44.2 & 54.5 & 44.1 & 53.0 & 53.9 & 52.2 & 46.7 & 47.4 \\
\cmidrule(r){2-3}\cmidrule(lr){4-4}\cmidrule(lr){5-5}\cmidrule(lr){6-6}\cmidrule(lr){7-7}\cmidrule(lr){8-8}\cmidrule(lr){9-9}\cmidrule(lr){10-10}\cmidrule(lr){11-11}\cmidrule(lr){12-12}\cmidrule(lr){13-13}

&\multirow{6}{*}{USAGE} 
&\blse   & 36.4 & 44.5 & 46.8 & 59.4 & 50.4 & 52.2 & 44.6 & 57.7 & 65.2 & 44.3 & 50.1\\
&&\muse & 13.0 & 31.5 & 34.9 & 63.3 & 43.0 & 17.4 & 35.2 & 25.7 & 56.9 & 31.2 & 35.2\\
&&\vecmap & 32.9 & 45.9 & 35.2 & 49.8 & 42.3 & 49.0 & 47.3 & \textbf{59.2} & 33.3 & 44.3 & 43.9\\ 
&&\mt     & 49.1 & 54.3 & 53.5 & 58.1 & 49.8 & 21.1 & 55.5 & 41.4 & 49.0 & 45.1 & 47.7 \\
&& \unsup & 56.8 & 48.33 & - & - & - & - & - & - & - & - & -\\
&& Ensemble & 48.2 & 55.5 & 42.7 & 57.1 & 50.4 & 28.9 & 53.3 & 48.4 & 44.5 & 48.0 & 47.7\\
\cmidrule(r){2-3}\cmidrule(lr){4-4}\cmidrule(lr){5-5}\cmidrule(lr){6-6}\cmidrule(lr){7-7}\cmidrule(lr){8-8}\cmidrule(lr){9-9}\cmidrule(lr){10-10}\cmidrule(lr){11-11}\cmidrule(lr){12-12}\cmidrule(lr){13-13}

&\multirow{6}{*}{SemEval} 
&\blse    & 31.6 & 55.3 & 37.9 & 47.9 & 56.4 & 70.3 & 58.3 & 43.4 & 44.5 & 47.9 & 49.3 \\
&&\muse & 18.2 & \textbf{69.8} & 56.6 & 53.6 & 63.8 & \textbf{87.5} & \textbf{59.9} & 36.7 & 50.0 & 57.2 & 55.3\\
&&\vecmap & \textbf{59.8} & 59.0 & 45.6 & 55.3 & \textbf{60.0} & 55.9 & 39.7 & 43.4 & 48.2 & 40.0 & 50.7 \\ 
&&\mt     & 57.0 & 58.7 & 40.5 & 58.2 & 49.0 & 61.6 & 57.6 & 40.3 & 53.8 & 50.8 & 52.8 \\
&& \unsup & 46.0 & 50.0 & - & - & - & - & - & - & - & - & -\\
&& Ensemble & 46.0 & 47.2 & 36.9 & 44.4 & 37.3 & 62.8 & 54.9 & 41.1 & 59.3 & 42.7 & 47.3 \\

\cmidrule(r){2-2}\cmidrule(r){3-3}\cmidrule(lr){4-4}\cmidrule(lr){5-5}\cmidrule(lr){6-6}\cmidrule(lr){7-7}\cmidrule(lr){8-8}\cmidrule(lr){9-9}\cmidrule(lr){10-10}\cmidrule(lr){11-11}\cmidrule(lr){12-12}\cmidrule(lr){13-13}\cmidrule(lr){14-14}

&& Average & 43.8 & 49.6 & 44.3 & 51.0 & 49.1 & 47.2 & 50.5 & 46.9 & 51.9 & 45.8 \\
\midrule

\multirow{19}{*}{\rt{Multi-class}} &
Training Data & Model & EU & CA & GL & IT & FR & NL & DE & NO & SV & DA & avg.\\
\cmidrule(r){2-2}\cmidrule(r){3-3}\cmidrule(lr){4-4}\cmidrule(lr){5-5}\cmidrule(lr){6-6}\cmidrule(lr){7-7}\cmidrule(lr){8-8}\cmidrule(lr){9-9}\cmidrule(lr){10-10}\cmidrule(lr){11-11}\cmidrule(lr){12-12}\cmidrule(lr){13-13}
& & maj. class  & 26.6 & 23.7 & 24.3 & 21.1 & 23.5 & 23.9 & 22.5 & 26.1 & 24.6 & 25.2 & 24.1\\
\cmidrule(r){3-3}\cmidrule(lr){4-4}\cmidrule(lr){5-5}\cmidrule(lr){6-6}\cmidrule(lr){7-7}\cmidrule(lr){8-8}\cmidrule(lr){9-9}\cmidrule(lr){10-10}\cmidrule(lr){11-11}\cmidrule(lr){12-12}\cmidrule(lr){13-13}
&\multirow{6}{*}{Twitter} 
&\blse & 32.6 & \textbf{35.9} & 30.1 & 26.7 & 28.0 & 28.7 & 36.9 & 41.4 & 40.9 & 24.3 & 32.6\\
&&\muse & 28.3 & 24.4 & 31.2 & 22.2 & 29.4 & 23.9 & 22.5 & 26.1 & 26.7 & 26.1 & 26.1\\
&&\vecmap & 26.5 & 30.2 & \textbf{39.6} & 26.7 & \textbf{37.2} & 34.6 & \textbf{39.8} & 31.7 & 33.4 & \textbf{41.0} & 34.1\\ 
&&\mt & 37.3 & 34.1 & 33.9 & 35.6 & 35.6 & \textbf{35.9} & 32.5 & \textbf{43.2} & 38.6 & 39.6 & \textbf{36.6} \\
&& \unsup & 40.1 & 28.5 & - & - & - & - & - & - & - & - & -\\
&& Ensemble & \textbf{41.5} & 30.5 & 36.5 & 26.9 & 36.3 & 31.9 & 30.9 & 37.9 & \textbf{42.8} & 36.3 & 35.1 \\
\cmidrule(r){2-3}\cmidrule(lr){4-4}\cmidrule(lr){5-5}\cmidrule(lr){6-6}\cmidrule(lr){7-7}\cmidrule(lr){8-8}\cmidrule(lr){9-9}\cmidrule(lr){10-10}\cmidrule(lr){11-11}\cmidrule(lr){12-12}\cmidrule(lr){13-13}

&\multirow{6}{*}{USAGE} 
&\blse      & 11.9 & 15.2 & 21.4 & 31.4 & 22.5 & 20.1 & 18.9 & 23.2 & 22.0 & 14.6 & 20.1 \\
&&\muse     & 3.3  & 21.1 & 15.6 & 28.9 & 18.9 & 5.9  & 18.6 & 14.3 & 24.2 & 16.8 & 16.8 \\
&&\vecmap   & 14.6 & 17.4 & 13.7 & \textbf{49.3} & 17.0 & 20.1 & 20.0 & 12.3 & 13.7 & 24.2 & 20.2 \\ 
&&\mt       & 19.8 & 22.5 & 23.9 & 26.2 & 18.3 & 10.2 & 24.8 & 19.4 & 16.1 & 13.2 & 19.4 \\
&& \unsup & 18.9 & 21.5 & - & - & - & - & - & - & - & - & -\\
&& Ensemble & 16.4 & 20.9 & 18.6 & 27.4 & 19.9 & 11.7 & 24.4 & 22.6 & 23.4 & 15.5 & 20.1 \\
\cmidrule(r){2-3}\cmidrule(lr){4-4}\cmidrule(lr){5-5}\cmidrule(lr){6-6}\cmidrule(lr){7-7}\cmidrule(lr){8-8}\cmidrule(lr){9-9}\cmidrule(lr){10-10}\cmidrule(lr){11-11}\cmidrule(lr){12-12}\cmidrule(lr){13-13}

&\multirow{6}{*}{SemEval} 
&\blse    &   13.6 & 24.9 & 13.8 & 20.4 & 24.9 & 26.9 & 24.6 & 18.5 & 18.7 & 19.2 & 20.6 \\
&&\muse   &   9.5  & 28.9 & 21.1 & 25.6 & 25.2 & 21.2 & 25.2 & 17.9 & 17.8 & 20.5 & 21.3 \\
&&\vecmap &   14.7 & 25.6 & 13.8 & 31.6 & 22.7 & 17.2 & 16.7 & 22.3 & 40.8 & 14.3 & 22.0 \\
&&\mt     &   15.2 & 24.0 & 19.1 & 26.2 & 20.0 & 25.1 & 26.8 & 20.6 & 19.2 & 15.7 & 21.2 \\
&& \unsup & 15.1 &  17.9 & - & - & - & - & - & - & - & - & -\\
&& Ensemble & 14.9 & 15.5 & 13.8 & 15.9 & 16.3 & 19.8 & 20.3 & 17.1 & 15.5 & 21.0 & 17.0 \\

\cmidrule(r){2-2}\cmidrule(r){3-3}\cmidrule(lr){4-4}\cmidrule(lr){5-5}\cmidrule(lr){6-6}\cmidrule(lr){7-7}\cmidrule(lr){8-8}\cmidrule(lr){9-9}\cmidrule(lr){10-10}\cmidrule(lr){11-11}\cmidrule(lr){12-12}\cmidrule(lr){13-13}\cmidrule(lr){14-14}

&& Average & 21.1 & 24.4 & 23.2 & 27.6 & 24.7 & 22.3 & 25.3 & 24.7 & 26.2 & 23.0 \\

\bottomrule
\end{tabular}
\caption[Deployment results]{Macro \F of targeted cross-lingual models
  on Twitter data in 10 target languages. Twitter refers to models
  that have been trained on the English data mentioned in Table
  \ref{deployment:table:tweets}, while USAGE and SemEval are trained
  on the English data from the datasets mentioned in Section
  \ref{targeteddatasets}.}
\label{deployment:table:results}
\end{table}
\end{small}

Table \ref{deployment:table:results} shows the macro \F scores for all
cross-lingual targeted sentiment approaches (\blse, \muse, \vecmap,
\mt) trained on English data and tested on the target-language using
the \splitsent method proposed in \ref{targetedprojection}. The final
column is the average over all languages. Given the results from the
earlier experiments, we hypothesize that \mt should outperform \muse,
\vecmap and \blse for most of the languages.

On the binary setup, \blse outperforms all other cross-lingual methods
including \mt and \unsup, with 56.0 macro averaged \F across languages versus
48.7, 49.4, and 48.9 for \muse, \vecmap, and \mt respectively (54.1 across Basque and Catalan versus 46.0 for \unsup). \blse
performs particularly well on Catalan (54.5), Italian (63.4), Swedish
(65.3), and Danish (68.3). \vecmap performs poorly on Galician (33.3),
Italian (38.2), and Danish (43.4), but outperforms all other methods
on Basque (56.4), Dutch (55.2) and Norwegian (59.0). \mt performs
worse than \blse and \vecmap, although it does perform best for
Galician (56.5). Unlike experiments in Section \ref{section:blse}, the
ensemble approach does not perform better than the individual
classifiers and \muse leads to the classifier with the lowest
performance overall. \unsup performs better than \mt on both Basque and Catalan.

On the multiclass setup, however, \mt (36.6 \F) is the best, followed
by \vecmap (34.1), \blse (32.6), and \muse (26.1). Compared to the
experiments on hotel reviews, the average differences between models
is small (2.5 percentage points between \mt and \vecmap, and 1.5
between \vecmap and \blse). \unsup performs better than \mt on Basque (40.1), but worse on Catalan (28.5). Again, all methods outperform the majority
baseline.

On both the binary and multiclass setups, the best overall results are
obtained by testing and training on data from the same domain (56.0 \F
for \blse and 36.6 \F for \mt). Training \mt, \muse, and \vecmap on
the SemEval data performs better than training on USAGE, however.

An initial error analysis shows that all models suffer greatly on the
negative class. This seems to suggest that negative polarity
towards a target is more difficult to determine within these
frameworks. A significant amount of the tweets that have negative
polarity towards a target also express positive or neutral sentiment
towards other targets. The averaging approach to create the context
vectors does not currently allow any of the models to exclude this
information, leading to poor performance on these instances.

Finally, compared to the experiments performed on hotel and product
reviews in Section \ref{sec:experiments}, the noisy data from Twitter
is more difficult to classify. Despite the rather strong majority
baseline (an average of 40.5 Macro \F on binary), no model achieves 
more than an average of 56 Macro \F on
the binary task. A marked difference is that \blse and \vecmap outperform \mt on the
binary setup. Unlike the previous experiment, \muse performs
the worst on the multiclass setup. The other projection methods obtain
multiclass results similar to the previous experiment (32.6--34.1 \F here compared
to 23.7--31.0 \F previously).

\subsection{Discussion}
\label{deployment:erroranlysis}

In this section, we present an error analysis. Specifically, Table \ref{helpful_examples} shows examples where \blse correctly predicts the polarity of a tweet that \mt and \unsup incorrectly predict, and vice versa, as well as examples where all models are incorrect.

In general, in examples where \blse outperforms \mt and \unsup, the translation-based approaches often mistranslate important sentiment words, which leads to prediction errors. In the first Basque tweet, for example, ``\#txindoki igo gabe ere inguruaz goza daiteke...	zuek joan tontorrera eta utzi arraroei gure kasa...'', \unsup incorrectly translates the most important sentiment word in the tweet ``goza'' (\textit{enjoy}) to ``overlook'' and subsequently incorrectly predicts that the polarity towards txindoki is negative.

Tweets that contain many out-of-vocabulary words or non-standard spelling (due to dialectal differences, informal writing, etc.), such as the third tweet in Table \ref{helpful_examples}, ``kanpora jun barik ehko asko: anboto, txindoki'', are challenging for all models. In this example ``jun'' is a non-standard spelling of ``joan'' (\textit{go}), ``barik'' is a Bizcayan Basque variant of ``gabe'' (\textit{without}) , and ``ehko'' is an abbreviation of ``Euskal Herriko'' (\textit{Basque Country's}). These lead to poor translations for \mt and \unsup, but pose a similar out-of-vocabulary problem for \blse.

\begin{figure}
\centering
\begin{minipage}{.5\textwidth}
  \centering
  \includegraphics[width=1\linewidth]{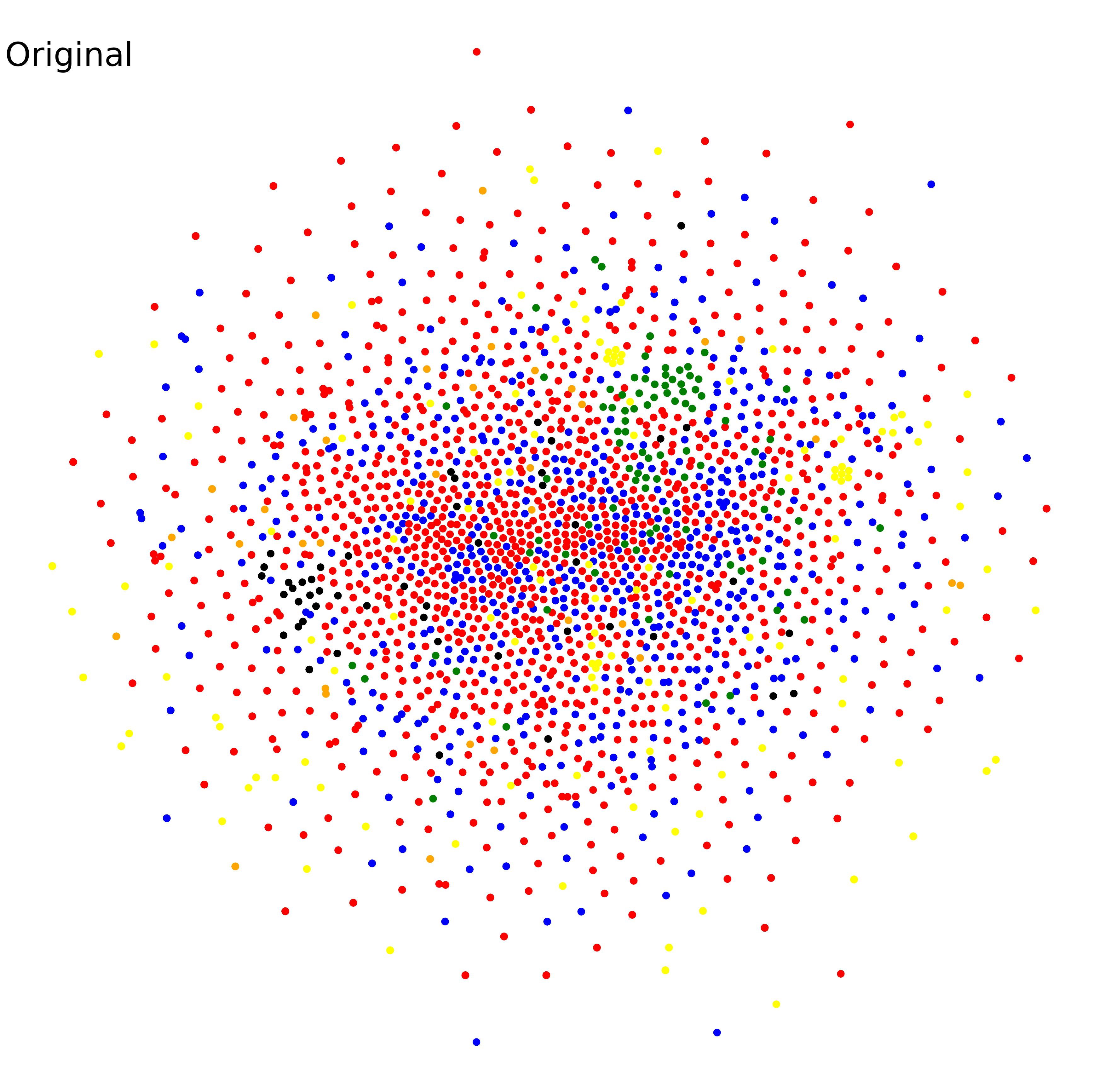}
  \label{fig:test1}
\end{minipage}%
\begin{minipage}{.5\textwidth}
  \centering
  \includegraphics[width=1\linewidth]{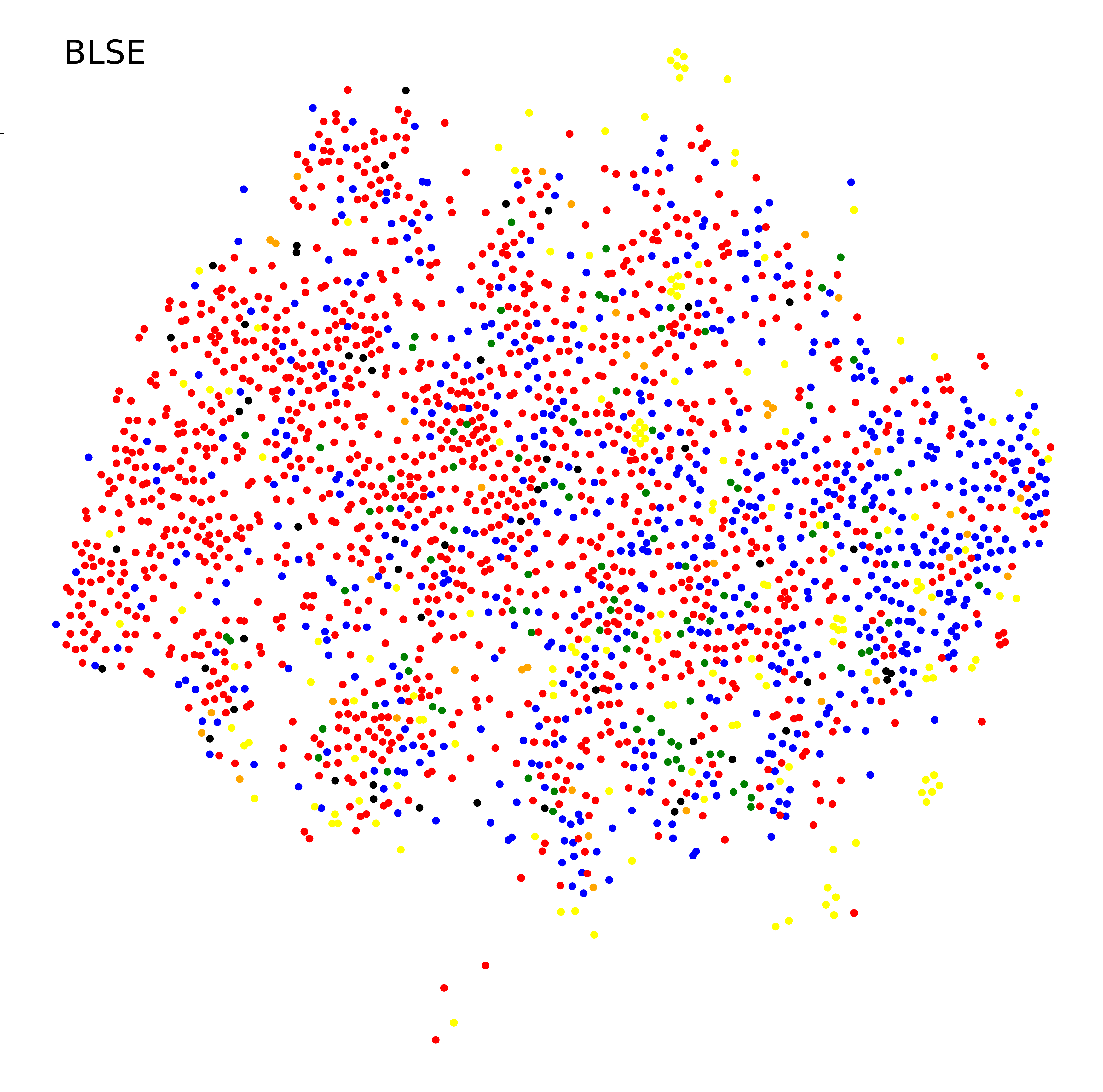}
  \label{fig:test2}
\end{minipage}
\caption[t-SNE visualization of \blse]{t-SNE-based visualization of the Basque vector space before
  and after projection with the targeted \blse. The positive and negative sentiment words are separated, although it is less clearly defined at target-level.}
\label{deployment:fig:tsne}
\end{figure}

In order to give a more qualitative view of the targeted model, Figure \ref{deployment:fig:tsne} shows t-sne projections of the bilingual vector space before and after training on the Basque binary task, following the same proceedure mentioned in Section \ref{qualitativeanalysis}. As in the sentence-level experiment, there is a separation of the positive and negative sentiment words, although it is less clear for targeted sentiment. This is not surprising, as a targeted model must learn not only the prior polarity of words, but how they interact with targets, leading to a more context-dependent representation of sentiment words.

%

Finally, we further analyze the effects of three variables 
that are present in cross-lingual sentiment analysis: a) availability
of monolingual unlabeled data, b) similarity of source and target languages,
and c) domain shift between the source language training data and
the target language test data.

\begin{small}
\begin{table}
\centering
\newcommand{\sepp}{\cmidrule(rl){1-1}\cmidrule(rl){2-2}\cmidrule(rl){3-3}}
\begin{tabular}{lll}
\toprule
Model & Tweet & Label \\
\sepp
\blse & \#txindoki igo gabe ere inguruaz goza daiteke...	& pos \\
 & zuek joan tontorrera eta utzi arraroei gure kasa... \\
\mt & \#txindoki it can also be enjoyed \underline{by the surrounding area}... & neg \\ 
& you go to the summit and leave it strange to us... \\
\unsup & \underline{Without falling also inguruaz overlook}... & neg \\
&you are rose summit and \underline{left arraroei on our own}... \\
Ref. & Even without climbing Txindoki, you can enjoy the surroundings... & \posbox{pos}\\
& go to the top and leave the weirdos to us...  \\
\sepp
\blse & ta gaur eiffel dorrea ikusiko degu, zelako gogoak aaaaaaaiiiiis!  & neg\\
\mt & \underline{ta} today we see eiffel tower, what kind of \underline{aaaaaaaiiiiis}! & pos \\
\unsup & \underline{i'm torn} eifell tower So we see the \underline{infamous enjoyment aaaaaaaiiiiis} & neg \\
Ref. & and today we'll visit the Eiffel Tower. So excited, ahhhhh! & \posbox{pos} \\
\sepp
\blse & kanpora jun barik ehko asko : anboto, txindoki ... & neu \\
\mt & \underline{many out of the jungle:} anboto & neu\\
\unsup & \underline{away jun at once congress on many : Shasta, Pine mountain} ... & neu\\
Ref. & Without leaving home, there's a lot of Basque Country: & \posbox{pos}\\
& Anboto, Txindoki ... \\
\sepp
\blse & é pisar a coruña e boom! venme unha onda de bandeiras españolas, & pos \\
& velliñxs falando castelán e señoras bordes e relambidas na cara. \\
\mt & is stepping on a coruña boom! I came across a wave of Spanish flags, & neg\\
& talking Spanish and \underline{ladies} and speaking on the face. PIC \\
Ref. & As soon as I get to a coruña, boom! A wave of Spanish flags, old people  & \negbox{neg}\\
& speaking Spanish, and rude, pretentious ladies hits me in the face. \\
\sepp
\blse & @musee louvre @parisotc purtroppo non ho visto il louvre.... & neg \\
& che file chilometriche !!! ( \\
\mt & @musee louve @parisotc unfortunately I did not see the louvre....  & pos \\
& \underline{that file kilometers!} ( \\
Ref. & @musee louve @parisotc unfortunately I did not see the louvre....  & \negbox{neg} \\
& the line was kilometers long! ( \\
\sepp
\blse & io voglio solamente andare al louvre :-( & neg \\
\mt & I just want to go to the louvre :-( & pos \\
Ref. & I just want to go to the louvre :-( & \posbox{pos} \\
\bottomrule
\end{tabular}
\caption{Examples where \blse is better and worse than \mt and \unsup. We show the original tweet in \blse, the automatic translation in \mt and \unsup, and reference translations (Ref.). The label column shows the prediction of each model and the reference gold label (either \posbox{pos} or \negbox{neg}). Additionally, we \underline{underline} relevant incorrect translations of words. }
\label{helpful_examples}
\end{table}
\end{small}

\subsubsection{Availability of Monolingual Unlabeled Data}

\begin{figure}[t]
  \centering
  \includegraphics[]{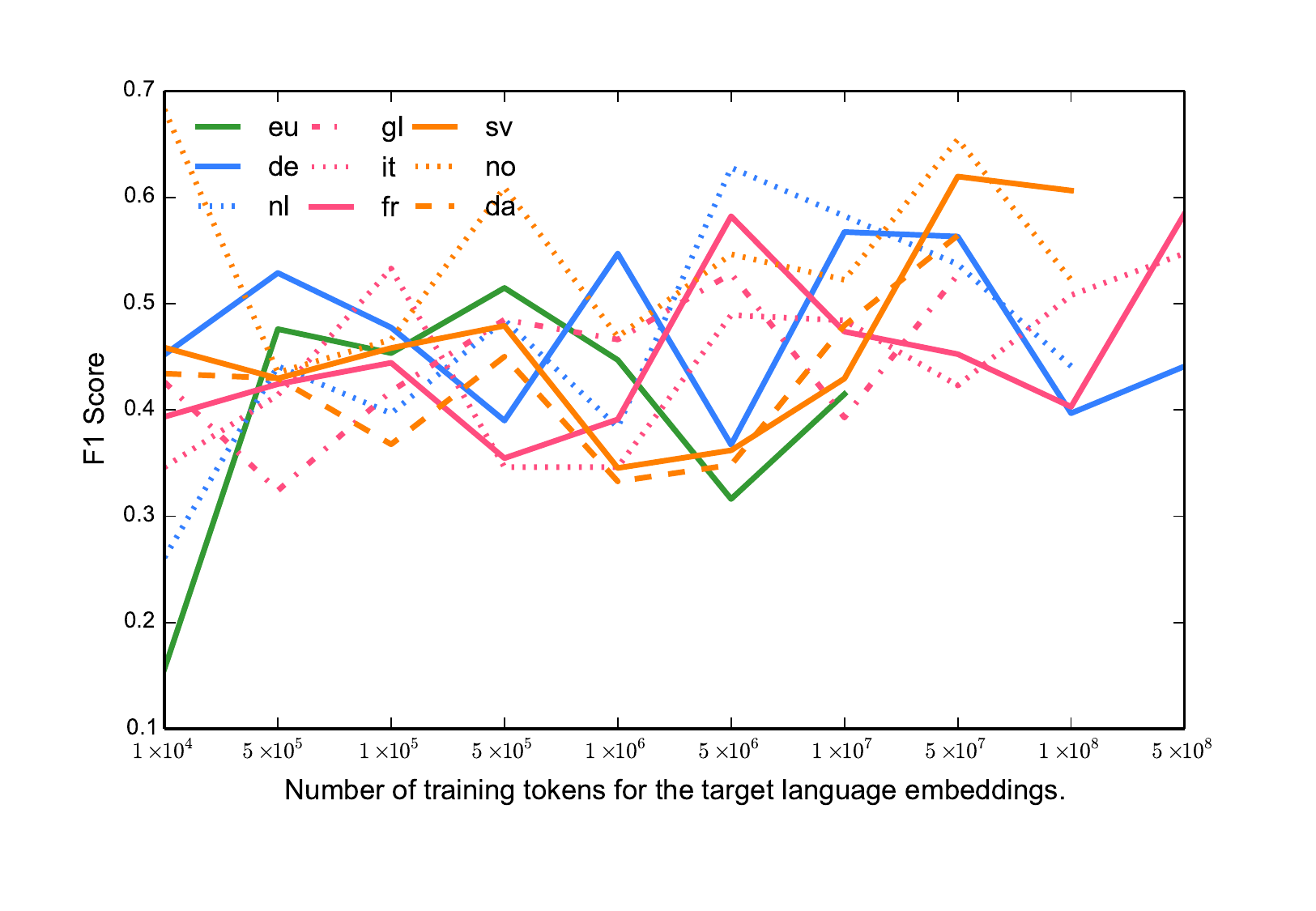}
  \caption[Results as a function of monolingual data]{Performance of
    \blse (Macro \F) on the binary sentiment task with training and
    test on Twitter as a function of amount of monolingual data
    available to train the monolingual embeddings in each
    language.}
  \label{deployment:fig:perfdata}
\end{figure}

We pose the question of what the relationship is between the amount of
available monolingual data to create the embedding spaces and the
classification results of the models. If the original word embedding
spaces are not of high quality, this could make it difficult for the
projection-based models to create useful features. In order to test
this, we perform ablation experiments by training target-language
embeddings on varying amounts of data ($1 \times 10^{4}$ to $5 \times
10^{9}$ tokens) and testing the models replacing the full
target-language embeddings with these. We plot the performance of the
models as a function of available monolingual data in Figure
\ref{deployment:fig:perfdata}.

Figure \ref{deployment:fig:perfdata} shows that nearly all models, with
the exception of Norwegian, perform poorly with very limited monolingual
training data ($1\times10^{4}$) and improve, although erratically, with
more training data. Interestingly, the models require little data to achieve results
comparable to using the all tokens to train the embeddings.  A
statistical analysis of the amount of unlabeled data available and the
performance of \blse, \muse, \vecmap (Pearson's $r$ =
$-0.14$, $-0.27$, $0.08$, respectively) reveals no
statistically significant correlation between them. This seems to 
indicate that all models are not sensitive to the amount
of monolingual training data available in the target language. 

\subsubsection{Language Similarity}
\label{deployment:langsimilarity}

\begin{figure}[t]
  \centering
  \includegraphics[width=1\textwidth]{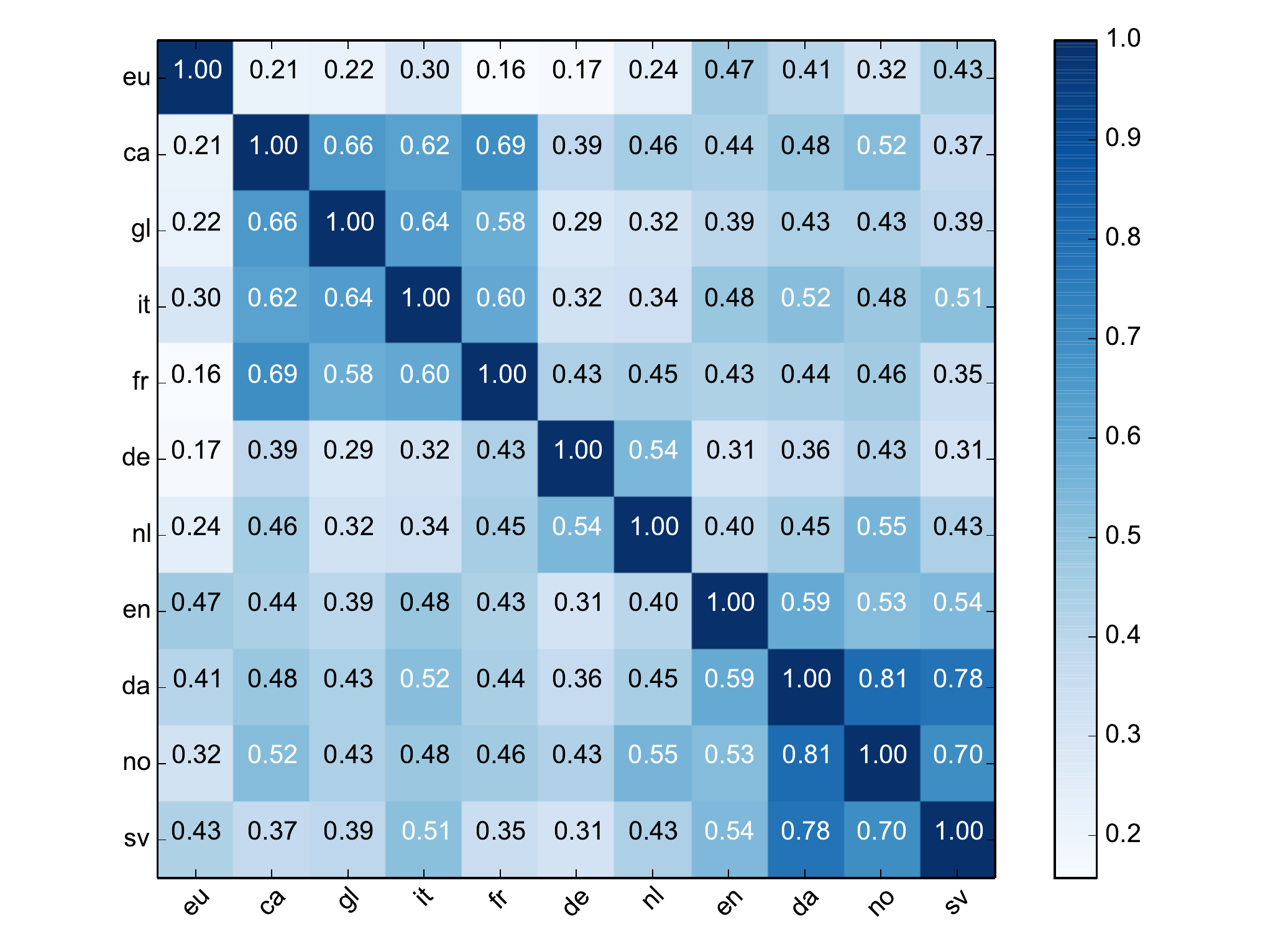}
  \caption[Language similarity]{Cosine similarity of 3-gram POS-tag
    and 3-gram character frequency.}
  \label{deployment:fig:similarity}
\end{figure}

One hypothesis to different results across languages is that the
similarity of the source and target language has an effect on the
final classification of the models. In order to analyze this, we need
a measure that models pairwise language similarity. Given that the
features we use for classification are derived from distributional
representations, we model similarity as a function of 1) universal
POS-tag n-grams which represent the contexts used during training, and
2) character n-grams, which represent differences in
morphology. POS-tag n-grams have previously been used to classify
genre \shortcite{Fang2010}, improve statistical machine translation
\shortcite{Lioma2005}, and the combination of POS-tag and character
n-grams have proven useful features for identifying the native
language of second language writers in English
\shortcite{Kulmizev2017}. This indicates that these are useful
features for characterizing a language. In this section we calculate
the pairwise similarity between all languages and then check whether
this correlates with performance.

\begin{figure}[t]
  \centering
  \includegraphics{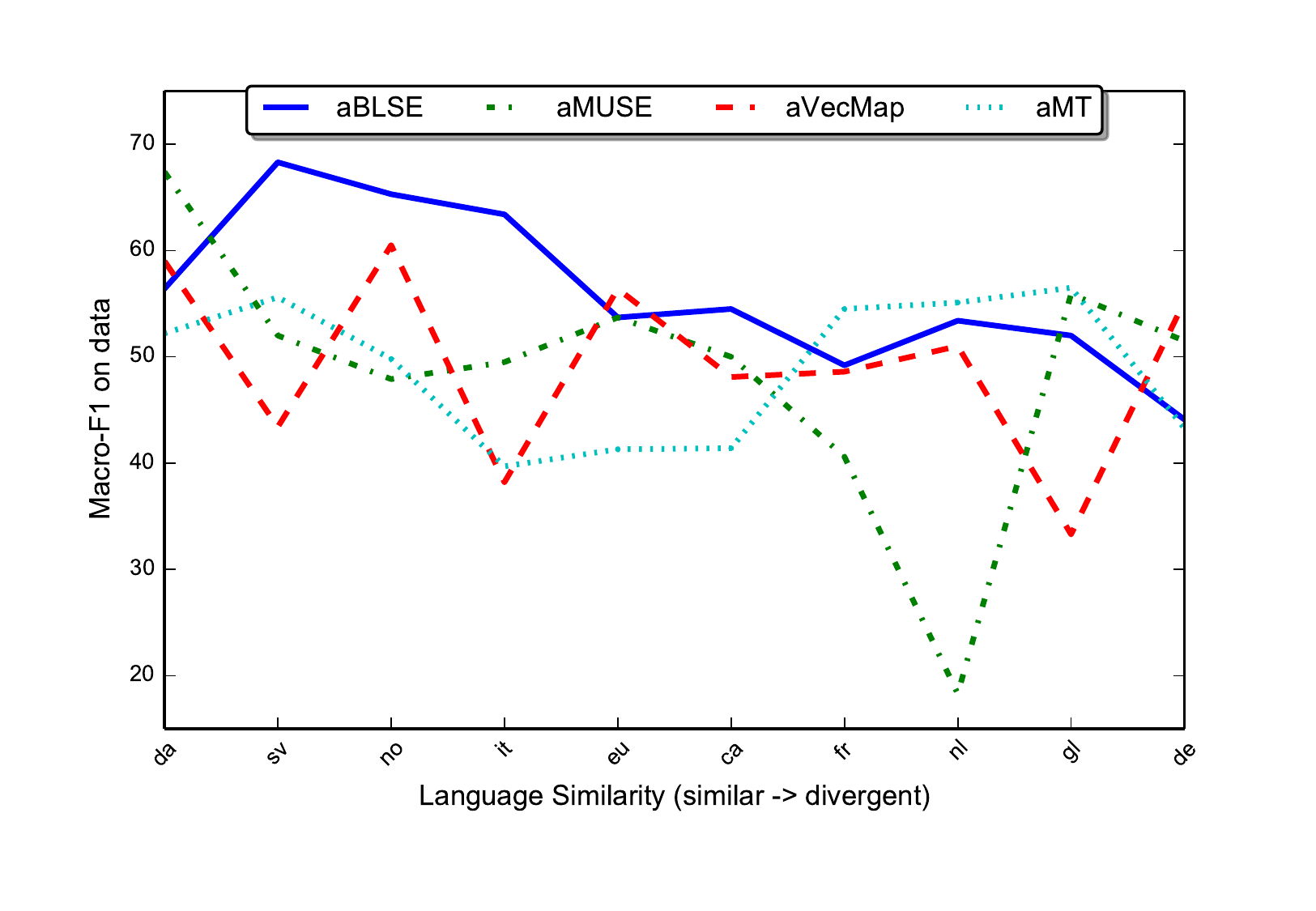}
  \caption[Results as a function of language similarity]{Performance
    (Macro \F) on the binary task as a function of cosine similarity
    between POS-tag and character trigram distributions in the source
    language (EN) and the target languages.}
  \label{deployment:fig:perfsimilarity}
\end{figure}

After POS-tagging the test sentences obtained from Twitter using the
universal part of speech tags \shortcite{Petrov2012}, we calculate the
normalized frequency distribution $P_{l}$ for the POS-tag trigrams and
$C_{l}$ for character trigrams for each language $l$ in $L =
\{\textrm{Danish, Swedish, Norwegian, Italian, Basque, Catalan,
  French, Dutch, Galician,}$ \\ $\textrm{German, English}\}$. We then
compute the pairwise cosine similarity between $\cos(A, B) = \frac{A
  \cdot B}{||A|| \: ||B||} $ where $A$ is the concatenation of
$P_{l_{i}}$ and $C_{l_{i}}$ for language $l_{i}$ and $B$ is the
concatenation of $P_{l_{j}}$ and $C_{l_{j}}$ for language $l_{j}$.

The pairwise similarities in Figure \ref{deployment:fig:similarity}
confirm to expected similarities, and language families are clearly
grouped (Romance, Germanic, Scandinavian, with Basque as an outlier
that has no more than 0.47 similarity with any language). This
confirms the use of our similarity metric for our purposes.
We plot model performance as a function of language similarity in
Figure \ref{deployment:fig:perfsimilarity}. To measure the correlation
between language similarity and performance, we calculate Pearson's
$r$ and find that for \blse there is a strong correlation between
language similarity and performance, $r = 0.76$ and significance $p <
0.01$. \muse, \vecmap and \mt do not show these correlations ($r$ =
0.41, 0.24, 0.14, respectively). For \mt this may be due to robust
machine translation available in less similar languages according to
our metric, \eg, German-English. For \muse and \vecmap, however, it is
less clear why it does not follow the same trend as \blse.

\subsubsection{Domain Similarity}

\begin{figure}[t]
  \centering
  \includegraphics[width=1\textwidth]{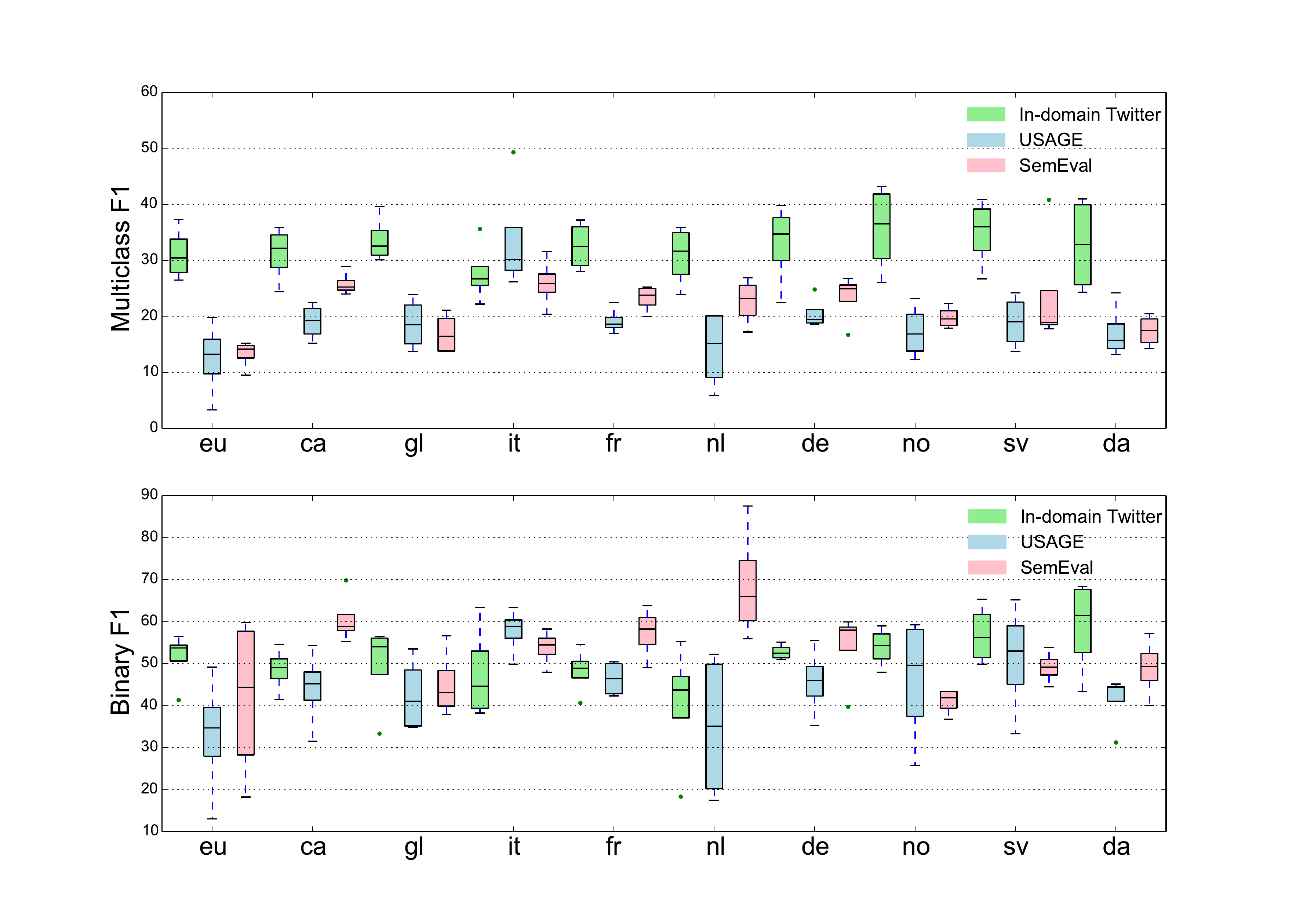}
  \caption[Domain effects]{Performance of all models (Macro \F) on the
    binary and multiclass task when trained on different source
    language data. For each target language, we show a boxplot for all
    models trained on \emph{In-domain Twitter data} (light green),
    \emph{USAGE product reviews} (light blue), and \emph{SemEval
      restaurant reviews} (pink). In the multiclass setup, we can see
    the in-domain data gives better results than the out-of-domain
    training data. This trend is not found in the binary setup,
    suggesting that binary classification is more robust to domain
    changes than multiclass classification.}
  \label{deployment:fig:domaindiffs}
\end{figure}

\begin{table}[]
\centering
\begin{tabular}{lccc}
\hline
         &   Twitter &   SemEval &   USAGE \\
\hline
 Twitter &     1.000 &     0.749 &   0.749 \\
 SemEval &     0.749 &     1.000 &   0.819 \\
 USAGE   &     0.749 &     0.819 &   1.000 \\
\hline
\end{tabular}
\caption[Domain similarity of English data]{Domain similarity of English training data measured as Jennson-Shannon divergence between the most common 10,000 unigrams.}
\label{deployment:table:domainsims}
\end{table}

In this section, we determine the effect of source-language domain on the cross-lingual sentiment classification task. Specifically, we use English language training data from three different domains (Twitter, restaurant reviews, and product reviews) to train the cross-lingual classifiers, and then test on the target-language Twitter data. In monolingual sentiment analysis, one would expect to see a drop when moving to more distant domains.

In order to analyze the effect of domain similarity further, we test the
similarity of the domains of the source-language training data using Jensen-Shannon Divergence, which
is a smoothed, symmetric version of the Kullback-Leibler Divergence,
$D_{KL}(A||B) = \sum_{i}^{N} a_{i} \log
\frac{a_{i}}{b_{i}}$. Kullback-Leibler Divergence measures the
difference between the probability distributions $A$ and $B$, but is
undefined for any event $a_{i} \in A$ with zero probability, which is
common in term distributions. Jensen-Shannon Divergence is then
\[
D_{JS}(A,B) = \frac{1}{2} \Big[ D_{KL}(A||B) + D_{KL}(B||A)  \Big]\,.
\]
Our similarity features are probability distributions over terms $t
\in \mathbb{R}^{|V|}$, where $t_{i}$ is the probability of the $i$-th
word in the vocabulary $V$. For each domain, we create frequency
distributions of the most frequent 10,000 unigrams that all domains
have in common and measure the divergence with $D_{JS}$.

The results shown in Table \ref{deployment:table:domainsims} indicate
that both the SemEval and USAGE datasets are relatively distinct from
the Twitter data described in Section \ref{deployment:datacollection},
while they are more similar to each other. Additionally, we plot the
results of all models with respect to the training domain in Figure
\ref{deployment:fig:domaindiffs}.

We calculate Pearson's $r$ on the correlation between domain and model
performance, shown in Table \ref{deployment:table:pearson_domain}. On
the binary setup, the results show a negligible correlation for \blse
(0.32), with no significant correlation for \muse, \vecmap or
\mt. This suggests that the models are relatively robust to domain
noise, or rather that there is so much other noise found in the
approaches that domain is less relevant. On the multiclass setup,
however, there is a significant effect for all models. This indicates
that the multiclass models presented here are less robust than the
binary models.

Both the SemEval and USAGE corpora differ equally from the Twitter data given the metric defined here. The fact that models trained on SemEval tend to perform better than those trained on USAGE, therefore, seems to be due to the differences in label distribution, rather than to differences in domain. These label distributions are radically different in the multiclass setup, as the English Twitter data has a 30/50/20 distribution over Positive, Neutral, and Negative labels (67/1/32 and 68/4/28 for USAGE and SemEval, respectively). Both undersampling and oversampling help, but the performance is still worse than training on in-domain data.

\begin{table*}[t]
\newcommand{\sepp}{\cmidrule(lr){2-2}\cmidrule(lr){3-3}\cmidrule(lr){4-4}\cmidrule(lr){5-5}\cmidrule(lr){6-6}}
 \centering
 \newcommand{\psta}{\phantom{*}}
 \renewcommand*{\arraystretch}{1.0}
 \setlength\tabcolsep{1.0mm}
\begin{tabular}{lccccc}
\toprule
 & \blse & \muse & \vecmap & \mt & \ensemble \\
\sepp
Binary & \psta$0.32$ & \psta$0.09$ & \psta$0.11$ & \psta$-0.07$ & \psta$-0.01$ \\
Multiclass & *$0.75$ & *$0.50$ & *$0.60$ & \phantom{$-$}*$0.88$ & \phantom{$-$}*$0.88$ \\
\bottomrule
\end{tabular}
\caption{Pearson's $r$ and $p$ values for correlations between domain
  and performance of each model. On the binary setup, there is no
  statistically significant effect of domain, while on the multiclass
  setup, all results are statistically significant ($p>0.01$, with Pearson's $r$ ).}
\label{deployment:table:pearson_domain}
\end{table*}

\subsubsection{Conclusion}
\label{deployment:discussion}

The case study which we presented in this section showed results of
deploying the models from Section \ref{sec:projecting} to real world
Twitter data, which we collect and annotate for targeted sentiment
analysis. The analysis of different phenomena revealed that for binary
targeted sentiment analysis, \blse performs better than machine
translation on noisy data from social media, although it is sensitive
to differences between source and target languages. Finally, there is
little correlation between performance on the cross-lingual sentiment
task and the amount of unlabeled monolingual data used to create the
original embeddings spaces which goes against our expectations.

Unlike the experiments in Section \ref{section:blse}, the ensemble
classifier employed here was not able to improve the results. We
assume that the small size of the datasets in this experiment does not
enable the classifier to learn which features are useful in certain
contexts.

One common problem that appears when performing targeted sentiment
analysis on noisy data from Twitter is that many of the targets of
interest are ambiguous, which leads to false positives. Even with
relatively unambiguous targets like ``Big Ben'', there are a number of
entities that can be referenced; Ben Rothlisberger (an American
football player), an English language school in Barcelona, and many
others. In order to deploy a full sentiment analysis system on Twitter
data, it will be necessary to disambiguate these mentions before
classifying the tweets, either as a preprocessing step or jointly.

In sentiment analysis, it is not yet common to test a model on
multiple languages, despite the fact that current state-of-the-art
models are often theoretically language-agnostic. This section shows
that good performance in one language does not guarantee that a model
transfers well to other languages, even given similar resources. We
hope that future work in sentiment analysis will make better use of
the available test datasets.

\section{Conclusion}
\label{sec:conclusion}
With this article, we have presented a novel projection-based approach
to targeted cross-lingual sentiment analysis. The central unit of the
proposed method is \blse which enables the transfer of annotations
from a source language to a non-annotated target language. The only
input it relies on are word embeddings (which can be trained without
manual labeling by self-annotation) and a comparably small translation
dictionary which connects the semantics of the source and the target
language.

In the binary classification setting (automatic labeling of sentences or
documents), \blse constitutes a novel state of the art on several
language and domain pairs. For a more fine-grained classification to
four sentiment labels, \barista and \muse perform slightly
better. The predictions in all settings are complementary to the strong upper
bound of employing machine translations: in an ensemble, even this
resource-intense approach is inferior.

The transfer from classification to target-level analysis revealed
additional challenges. The performance is lower, particularly for the
4-class setting. Our analyses show that mapping of sentence
predictions to the aspects mentioned in each sentence with a machine
translation model is a very challenging
empirical upper bound -- the difference in performance compared 
to projection-based methods is greater
here than for the sentence-classification setting.
However, we showed that in resource-scarce environments, \blse
constitutes the current state of the art for binary target-level sentiment
analysis when incorporated in a deep learning architecture which is
informed about the aspect. \muse performs better in the same
architecture for the 4-class setting.

Our analysis further showed that the neural network needs to be
informed about both the aspect and the context -- limiting the
information to a selection of these sentence parts strongly
underperforms the combined setting. That also demonstrates that the model
does not rely on prior distributions of aspect mentions.

The final experiment in the paper is a real-world deployment of the
target-level sentiment analysis system in multilingual setting with 10
languages, where the assumption is that the only supervision is
available in English (which is not part of the target languages). We
learned here that it is important to have access to in-domain data
(even for cross-lingual projection), especially in the multiclass
setting. Binary classification however, which might often be
sufficient for real-world applications, is more robust to domain
changes. Further, machine translation is less sensitive to language
dissimilarities, unlike projection-based methods. The
amount of available unlabeled data to create embeddings plays a role
in the final performance of the system, although only to a minor extent.

The current performance of the projection-based techniques still lags
behind state-of-the-art \mt approaches on most tasks, indicating that there is still much work to be done. While general bilingual embedding techniques do not seem to incorporate enough sentiment information, they are able to retain the semantics of their word vectors to a large degree even after projection. We hypothesize that the ability to retain the original semantics of the monolingual spaces leads to \muse performing better than \mt on multiclass targeted sentiment analysis. The joint approach introduced in this work suffers from the degradation of the original semantics space, while optimizing the sentiment information. Moving from a similarity-based loss to a ranking loss, where the model must predict a ranked list of most similar translations could improve the model, but would require further resource development cross-lingually, as a simple bilingual dictionary would not provide enough information.

One problem that arises when using bilingual embeddings instead of
machine translation is that differences in word order are no longer
handled \cite{Atrio2019}. Machine translation models, on the other hand, always include
a reordering element. Nonetheless, there is often a mismatch between
the real source language word order and the translated word order. In
this work, we avoided the problem by using a bag-of-embeddings
representation, but \shortciteA{Barnes2017} found
 that the bag-of-embeddings approach does not perform
as well as approaches that take word order into account, \eg, \lstms
or \cnns. We leave the incorporation of these
classifiers into our framework for future work.

Unsupervised machine translation
\shortcite{Artetxe2018,Lample2018,artetxe2018emnlp} shows great promise for sentence-level classification. Like \mt, however, it performs worse on noisy data, such as tweets. Therefore, users who want to apply targeted cross-lingual approaches to noisy data should consider currently consider using embedding projection methods, such as \blse. Future work on adapting unsupervised machine translation to noisy text may provide another solution for low-resource NLP.

\pagebreak

\acks{The authors thank Patrik Lambert, Toni Badia, Amaia Oliden, Itziar
Etxeberria, Jessie Kief, Iris Hübscher, and Arne Øhm for helping with the annotation of the resources used in this research. This
  work has been partially supported by the DFG Collaborative Research
  Centre SFB 732 and a SGR-DTCL Predoctoral Scholarship.}

\vskip 0.2in
\bibliography{lit}
\bibliographystyle{theapa}

\end{document}